\documentclass[10pt,journal,compsoc]{IEEEtran}
\usepackage[caption=false]{subfig}
\usepackage{adjustbox}
\usepackage{silence}
\usepackage{array}
\newcolumntype{x}[1]{>{\centering\arraybackslash\hspace{0pt}}p{#1}}
\usepackage{graphicx}
\usepackage[utf8]{inputenc} 
\usepackage[T1]{fontenc}    
\usepackage{hyperref}       
\usepackage{url}            
\usepackage{booktabs}       
\usepackage{amsfonts}       
\usepackage{nicefrac}       
\usepackage{microtype}      
\usepackage{xcolor}         
\usepackage{float}
\usepackage{algorithm}
\usepackage{algorithmic}
\usepackage{wrapfig}
\usepackage{supertabular,booktabs}

\usepackage{pythonhighlight}
\usepackage[center]{caption}
\usepackage{listings}
\usepackage{longtable}
\usepackage{amsmath,amscd,amsbsy,amssymb,latexsym,url,bm,amsthm}
\newtheorem*{remark}{Remark}

\newtheorem*{corollary}{Corollary} 
\usepackage{hyperref}
\usepackage{balance}

\ifCLASSOPTIONcompsoc
  \usepackage[nocompress]{cite}
\else
  \usepackage{cite}
\fi
\ifCLASSINFOpdf
\else
\fi
\begin{document}
%
\title{Curriculum-based Asymmetric Multi-task Reinforcement Learning}

%
%
%

\author{Hanchi~Huang,
        Deheng~Ye,
        Li~Shen, and Wei~Liu
\IEEEcompsocitemizethanks{
\IEEEcompsocthanksitem Hanchi Huang is with Nanyang Technological University, Singapore. Email: hhuang036@e.ntu.edu.sg.  Deheng Ye and Wei Liu are with Tencent Inc., China. Email: dericye@tencent.com, wl2223@columbia.edu. Li Shen is with JD.com Inc., China. Email: mathshenli@gmail.com
\IEEEcompsocthanksitem Deheng Ye and Wei Liu are the corresponding authors. 
}
}

\IEEEtitleabstractindextext{%
\begin{abstract}

We introduce CAMRL, the first curriculum-based asymmetric multi-task learning (AMTL) algorithm for dealing with multiple reinforcement learning (RL) tasks altogether. 
To mitigate the negative influence of customizing the one-off training order in curriculum-based AMTL, 
CAMRL switches its training mode between parallel single-task RL and asymmetric multi-task RL (MTRL), according to an indicator regarding the training time, the overall performance, and the performance gap among tasks.
To leverage the multi-sourced prior knowledge flexibly and to reduce negative transfer in AMTL, we customize a composite loss with multiple differentiable ranking functions and optimize the loss  through alternating  optimization and the Frank-Wolfe algorithm.
The uncertainty-based automatic adjustment of hyper-parameters is also applied to  eliminate the need of laborious hyper-parameter analysis during optimization.
By optimizing the composite loss, CAMRL predicts the next training task and continuously revisits the transfer matrix and network weights. 
We have conducted experiments on a wide range of benchmarks in multi-task RL, covering Gym-minigrid, Meta-world,  Atari video games,  vision-based PyBullet tasks, and RLBench, to show the improvements of CAMRL over the corresponding single-task RL algorithm and state-of-the-art MTRL algorithms. The code is available at: \url{https://github.com/huanghanchi/CAMRL}.

\end{abstract}

\begin{IEEEkeywords}
Reinforcement learning, curriculum learning, asymmetric multi-task learning. 
\end{IEEEkeywords}}

\maketitle
 
\IEEEdisplaynontitleabstractindextext 
%
\IEEEpeerreviewmaketitle

\IEEEraisesectionheading{\section{Introduction}\label{sec:introduction}}

%
%
%
%
\begin{figure*}[ht!]
    \centering
\includegraphics[width=0.7\linewidth]{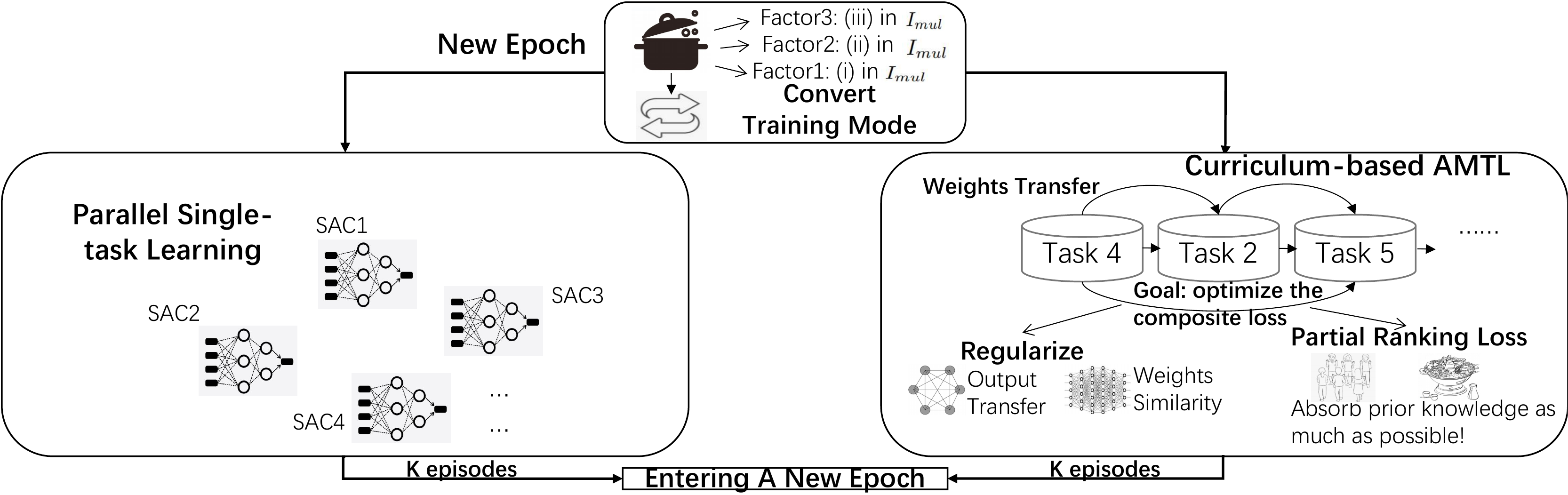}
    \caption{The workflow of CAMRL. When an epoch starts, CAMRL switches its training mode to single-task learning or curriculum-based AMTL. 
    For the latter mode, CAMRL customizes a curriculum for multiple tasks and updates the transfer graph  between tasks via a composite loss. 
   CAMRL keeps the training mode for $K$  episodes before entering a new epoch.}
    \label{fig:CAMRL Pipeline}
\end{figure*}

\IEEEPARstart{M}{ulti-task}
learning (MTL) trains multiple related tasks simultaneously while taking   advantage of similarities and differences between tasks \cite{ZhangY17aa}. 
However, in MTL, negative transfer may occur since not all tasks can benefit from joint learning. 
To handle this,  asymmetric  transfer between every pair of tasks has been developed \cite{lee2016asymmetric}, such that the amount of  transfers from a confident network to a relatively less confident one is larger than the other way around.  
The core idea is to learn a sparse weighted directed regularization graph between every two tasks through a curriculum-based asymmetric multi-task learning (AMTL) mode,
which has been proven to be very effective in supervised learning  \cite{pentina2015curriculum,lee2016asymmetric,lee2018deep}, while remaining to be unexplored in reinforcement learning (RL). 

In this paper, inspired by AMTL, we study the problem of mitigating negative transfer in multi-task reinforcement learning (MTRL).
Note that incorporating AMTL techniques into the context of RL is not easy.
Due to the non-stationarity when training  RL tasks and the lack of prior knowledge on RL tasks' properties, it is likely that we end up with learning a poor regularization graph between tasks, if we directly adopt curriculum-based AMTL to train tasks one by one without proper corrections to the training order. 
Besides, some important factors for flexiblizing the transfer among tasks and for avoiding negative transfer, such as indicators representing relative training progress, the transfer performance between tasks, and the behavioral diversity  described in \cite{portelas2020automatic}, have been neglected by existing works on AMTL \cite{pentina2015curriculum,lee2016asymmetric,lee2018deep}, which may lead to slow convergence for a RL task, and cause serious negative transfer before convergence.
Also, it is worth noting that performing curriculum-based AMTL all the time is time-consuming and can lead to performance degradation when each task is poorly trained or has already been well-trained. 
Consequently, RL tasks require different demands for curriculum learning at different training stages and the amount of experience learned from different tasks needs to be dynamically adjusted.

To deal with the above issues, we propose CAMRL, the first {\bf C}urriculum-based {\bf A}symmetric {\bf M}ulti-task algorithm for {\bf R}einforcement {\bf L}earning.
To balance both the efficiency and the amount of curriculum-based transfers across tasks, we design a composite indicator  weighting multiple factors 
to decide which training mode to switch   between parallel single-task learning and curriculum-based AMTL. 
To avoid customizing the one-off order of training tasks, CAMRL re-calculates the above indicator at the beginning of every epoch to switch the training mode. When entering the AMTL training mode,  CAMRL 
updates  the training order, the regularization graph, and the network weights by optimizing a composite loss function through  alternating optimization and the Frank-Wolfe algorithm. The loss function regularizes both the amount of outgoing transfers and the similarity across multiple network weights. Furthermore, three novel differentiable  ranking functions are proposed  to flexibly incorporate various prior knowledge  into the loss, such as
the relative training difficulty, the performance of mutual evaluation, and the similarity  between tasks. 
Finally, we discuss the flexibility of CAMRL from multiple perspectives and its limitations, so as to illuminate further incorporations of curriculum-based AMTL into RL.

Our contributions are summarized as follows: 

\begin{itemize}

\item 
We propose the CAMRL (curriculum-based asymmetric multi-task reinforcement learning) algorithm, which is designed with a training-mode-switching mechanism and a combinatorial loss containing multiple differentiable ranking functions.
These designs tackle the common issues in multi-task RL, e.g., negative transfer and poor utilization of prior knowledge.

 
\item CAMRL can be paired with various RL-based algorithms and training modes, as well as  absorbing various prior knowledge and ranking information of training factors of any amount, which are rarely seen in previous MTRL works \cite{teh2017distral,yu2020gradient,yang2020multi}.
Moreover, by dynamically adjusting parameters, CAMRL can adapt to the arrival of a new task via a simple correction.
 
\item Experiments on a series of low-/high- dimensional RL tasks show that CAMRL remarkably outperforms the corresponding single-task RL algorithm and state-of-the-art multi-task RL algorithms.  

\end{itemize}

\section{Related work}
\subsection{Multi-Task Reinforcement Learning (MTRL)}
Before stepping into the deep RL era, most  
works \cite{boutsioukis2011transfer,brunskill2013sample}  on  multi-task-oriented algorithms within RL
attempted to rely on the assistance from transfer learning. 
Later on, knowledge sharing in multi-task deep RL was proposed to distill the common traits among all tasks \cite{teh2017distral,espeholt2018impala}.  
  Since negative  interference  among gradients from various tasks may occur very often during the distilling process, the work \cite{yu2020gradient} presented a  gradient surgery method that alters
gradients by projecting each onto the normal plane of the other when conflicts happen between gradients.  Researchers also tried to capture the similarity between gradients from multiple tasks \cite{zhang2014regularization,chen2018gradnorm}. 
 Furthermore, apart from   driving  gradients to be similar,  applying compositional models is also a natural approach to handle  gradients' conflicts. 
By breaking down a  multi-task problem into modules and supporting composability of separate modules in new modular RL agents, the  compositional model enables to relieve the conflicts among gradients and facilitate better generalization \cite{devin2017learning}. 
 However, training sub-policies separately often requires well-predefined subtasks, which may be infeasible in real-world applications. 
Rather than manually defining the modules or sub-policies and the specification of their combination, the authors of \cite{yang2020multi} introduced the  soft module, a fully end-to-end method    which generates soft combinations of multiple modules automatically without specifying the policy structure in advance.

\subsection{Curriculum Learning}
Curriculum learning (CL) describes a learning mode in which we start with  easy tasks and then gradually increase the task difficulty.
The idea of using curricula to train learning agents was first proposed in \cite{elman1993learning} and first introduced for deep learning by \cite{bengio2009curriculum}. 
Over the years, curricula have been applied to train complex tasks  or tasks with sparse rewards \cite{asada1996purposive,wu2016training,bansal2017emergent,ye_nips_2020}. 
Early research efforts focused on the manual configuration of the curriculum which can be time-consuming and requires domain knowledge. 
To deal with the above issue, automatic CL has gained popularity recently, which aims to automatically adapt the distribution of training data by learning to adjust the selection of curricula according to some factors such as diversity and learning progress \cite{portelas2020automatic}.  Our work is, however, based on  another line of works \cite{pentina2015curriculum,lee2016asymmetric} that select the next task to train by performing joint optimization on the loss of supervised learning tasks with $\ell_2$-norm regularizations on task parameters.

\section{Methodology}\label{algorithm}

\subsection{Overview}

The pipeline of CAMRL is shown in Figure \ref{fig:CAMRL Pipeline}.
In CAMRL, we customize a learning progress indicator to determine whether to perform parallel single-task training  or curriculum-based  AMTL at the beginning of each epoch. When performing  curriculum-based  AMTL, 
a composite loss function that learns the transfer between  tasks and considers several factors to avoid negative training is adopted. 
We apply the alternating optimization and Frank-Wolfe algorithm to decide the next task for training and update the transfer matrix. 
In the meantime,
the hyper-parameters in the loss are automatically adjusted according to each term's historical uncertainty.
When a new task arrives, CAMRL can quickly adapt to the new scenario without much side effect to the original tasks, by merely modifying the transfer matrix.

\begin{algorithm*}[thb!]
\flushleft
\begin{algorithmic}[1]
	\STATE \textbf{Inputs: } $\mu_1,\mu_2,\lambda_i(i=0,1,2,3,4)$,  number of tasks  $T$.  
	\STATE 
      \textbf{Initialization:} The transfer matrix $B=I^{T\times T}$; the soft actor-critic network $SAC_t$ with parameters $w_t$ for $t\in [T]$; $W=(w_{1}, w_{2}, \ldots, w_{T})$.
   
       \FOR{\texttt{$n=1,2,\cdots,N$}}
        \STATE Perform parallel training of all $SAC_t$ for $t\in [T]$ for $K$  episodes without interference among losses.
      \ENDFOR  
      
       \FOR{\texttt{$n=N+1,N+2,\cdots$}}
       \STATE 
       Calculate $I_{mul}$ and sample $u\sim Uniform([0,1])$. 
       \IF{$u<I_{mul}$}
     \STATE   Perform parallel training of $SAC_t (t\in [T])$  for $K$ episodes without interference among losses. \label{line:11}
       \ELSE
        \STATE Predict the next training task $t$ and update $b_{t}^{o}=(B_{t 1}, \ldots, B_{t(t-1)}, B_{t(t+1)},\cdots,B_{tT})$ and $w_t$ by optimizing:
\begin{align}
&(t, b_{t}^{o})\! \leftarrow \!\underset{t \in \mathcal{U}, b_{t}^{o}}{\arg\min}\Big\{\lambda_0[(1\!+\!\mu_1\|b_{t}^{o}\|_{1}) \mathcal{L}(w_{t} ) \!-\!\mu_2 (b_t^o)^\top l_t^o] \!+\!\lambda_1\!\!\! \sum_{s \in \mathcal{U}  -  t}\!\!\Big\|w_{s}\!-\!\!\sum_{j=1}^{i-1}\!\! B_{\pi(j) s} w_{\pi(j)}\!-\!B_{t s} w_{t}\Big\|_{2}^{2}  \nonumber \\
&\qquad\quad \!+\!\lambda_2 \sum_{j\in [q]}(j-y{\prime}_{i_j})^2+\lambda_3 \sum_{j\in [T]}(rank^{(1)}_j-y{\prime\prime}_{j})^2  +\lambda_4 \sum_{j\in [T]}(rank^{(2)}_j-y{\prime\prime}_{j})^2
\Big\}.\label{eq:7}
\end{align}
\STATE  
Fix $b_{t}^{o},W$ and select the task $t$ which minimizes the objective in Eq. \eqref{eq:7}. 
\STATE
Fix $t,W$ and run vanilla Frank-Wolfe on  Eq. \eqref{eq:6} for several iterations  to optimize $b_{t}^{o}$.
\STATE Fix $b_{t}^{o}$ and train task $t$  with the policy loss Eq. \eqref{eq:7} for $K$ episodes.
\ENDIF
\ENDFOR  
\end{algorithmic}
\caption{CAMRL Algorithm} 
\label{alg:CAMRL}
\end{algorithm*}

The overall training procedure of CAMRL is summarized in Algorithm \ref{alg:CAMRL}. Assume that we have $T$ tasks with varying degrees of difficulty. In the beginning, we simply equip each task $t\in [T]$ with a soft actor-critic (SAC) network $SAC_t$ \cite{HaarnojaZAL18}, and perform parallel single-task training for several epochs without interference among losses for different tasks, that is, each $SAC_t (t\in [T])$ is trained independently. Denote $w_t$ as the network parameters of the $t$-th SAC network.  Next, according to an indicator related to the learning progress, CAMRL decides whether to switch the training mode into curriculum-based  AMTL, which trains the tasks one by one by optimizing a composite loss term regularizing the transfer matrix between tasks. The indicator $I_{mul}(=\frac{(i)+(ii)+(iii)}{3})$ is the positive weighted sum  of the following terms: (i) $\exp(-n/a)$, where $n$ refers to the number of the current epochs. We use $\exp(-n/a)$ to encourage  larger probability of multi-task training  in the early stage so that hard-to-train  tasks   could benefit from easy-to-train tasks as early as possible. As we will show in Table \ref{tab:ablation study}, the performance of CAMRL  downgrades obviously when we remove this term in $I_{mul}$; 
(ii) $\exp(-\mathcal{L}_{nor}*b)$, where $\mathcal{L}_{nor}$ means the average normalized 
policy loss of all tasks during the last epoch. We use $\mathcal{L}_{nor}$ to judge the overall training progress. If the overall training performance is poor, i.e. with a large $\mathcal{L}_{nor}$, then the possibility of performing inter-task transfer tends to be smaller. Note that in experiments, we set $\mathcal{L}_{nor}$ to be $\frac{policy\_loss-average_{policy\_loss}}{std_{policy\_loss}}$.  See the following remark for the details of the policy loss; 
(iii) the percentage of tasks whose average normalized reward during the last epoch locates outside the interval $[\mathcal{R}_{nor}-c*std_{nor},\mathcal{R}_{nor}+c*std_{nor}]$, where $std_{nor}$ is the standard deviation of the normalized rewards for all tasks during that period. This term indicates that the larger the learning process gap among tasks, the larger the probability of performing  multi-task training.

\begin{remark}
 \textbf{Policy loss of SAC in \cite{yang2020multi}.}  Due  to the implementation issue of SAC inheriting from the code of the soft-module \cite{yang2020multi}, we find it much more convenient to use the negative normalized version of the policy loss, the direct output of the 'train' function in the code of the soft-module,
instead of the normalized return. Besides, when
encountering into difficult tasks with sparse returns,  the normalized return might keep nearly unchanged from time to time. In such scenarios, utilizing the negative normalized
policy loss can help speed up model training more than  the normalized return.
Specifically, the policy loss of the SAC used in \cite{yang2020multi} can be characterized as $\log \pi_{\theta}\left(a_{t} \mid s_{t}\right)-Q_{\theta}\left(a_{t}, s_{t}\right)$, where $\pi_{\theta}\left(a_{t} \mid s_{t}\right)$ is the probability of adopting action $a_t$ in face of the state $s_t$ and $Q_{\theta}\left(a_{t}, s_{t}\right)$ is the estimated value of 
the state-action pair $(a_t, s_t)$.

\end{remark}

If $u\sim Uniform([0,1])$ is smaller than $I_{mul}$, then the curriculum-based AMTL training mode will be temporarily adopted in the next epoch; otherwise, we will perform the single-task training mode (see line 8-10  in Algorithm  \ref{alg:CAMRL}).
Here we need to note that: $I_{mul}$ is constantly changing, it does not matter if its value is temporarily greater than 1, which indicates that performing curriculum-based AMTL training is necessary at this time.  When curriculum-based AMTL is unnecessary, $I_{mul}$  falls back down again and becomes smaller than 1.

\subsection{Curriculum Multi-task Training}
In this subsection, we follow \cite{lee2016asymmetric} to establish the foundation of curriculum multi-task training for our CMARL. 

Let $B$ be a $T \times T$  matrix symbolizing the count of  transfers between each pair of tasks. 
For  $t \in [T]$, let $w_t$ be the parameters of the $t$-th soft actor-critic network.
We follow the assumption in  \cite{lee2016asymmetric} that  for all $ t \in [T], w_{t} \approx \sum_{s=1}^{T} B_{s t} w_{s}$, that is, $B_{st}$  refers to the positive weight of basis $w_s$  representing $w_{t}$.  Denote $W:=(w_{1}, w_{2}, \ldots, w_{T})$.
  
The core of CAMRL is adapted from the following composite loss \cite{lee2016asymmetric}:
\begin{align}
\mathcal{L}(W,B)=\sum_{t=1}^{T}\{(1+\mu\|b_{t}^{o}\|_{1}) \mathcal{L}(w_t) +\lambda\|w_{t}-\sum_{s \neq t} B_{s t} w_{s}\|_{2}^{2}\}, \label{eq:1} 
\end{align}
where  $b_{t}^{o}\!=\!(B_{t 1}, \ldots, B_{t(t-1)}, B_{t(t+1)}, \ldots, B_{t T})^{\top}\!\in\!\mathbb{R}^{T-1}$  represents the count of outgoing transfers from task $t$ to other tasks and we use $\|b_{t}^{o}\|_{1}$  to satisfy the sparsity property of the transfer matrix $B$;
$\mathcal{L} (w_t)$ is the  policy loss for task $t$ under $w_t$, and $w_t$ is the parameters of the $t$-th SAC network; 
$(\lambda, \mu)$ are the coefficients of weights for different terms in Eq.~\eqref{eq:1}. 

Since simultaneous training of $W$ and $B$ may result in a
serious negative transfer and a dimensional disaster when performing optimization, Lee et al. \cite{lee2016asymmetric}  formulated the curriculum-based training mode
leveraging the paradigm in Eq.~\eqref{eq:1}, for the purpose of finding the optimal order of training tasks and only optimizing $b_t^o$ and $w_t$ rather than $B$ and $W$  when training  the task $t$.

Denote $\mathcal{S}$ as the permutation space over $T$ elements, and for  $\pi \in \mathcal{S}$, denote $\pi(i)$ as the  $i$-th element in permutation  $\pi$. 
To perform curriculum learning of multiple tasks, Lee et al.  \cite{lee2016asymmetric} adapted their
goal  to cope with Eq. \eqref{eq:2}:
\begin{align}
    \underset{\pi \in \mathcal{S}, W, B \geq 0}{\min}  &\sum_{i=1}^{T}\Big\{(1+\mu\|b_{\pi(i)}^{o}\|_{1})  \mathcal{L}(w_{\pi(i)} )\nonumber   \\& +\lambda\|w_{\pi(i)}-\sum_{j=1}^{i-1} B_{\pi(j) \pi(i)} w_{\pi(j)}\|_{2}^{2}\Big\}.  \label{eq:2} 
\end{align}

Let  $\mathcal{T}:=\{\pi(1), \ldots, \pi(i-1)\}$ be  trained tasks and $\mathcal{U}=\{1, \ldots, T\}  -  \mathcal{T}$ be  untrained tasks  in the current epoch, respectively. Lee et al.  \cite{lee2016asymmetric} then found a task $t\in U$ to be
learned next, so as to improve the future learning process the most: 
\begin{align}
(t, b_{t}^{o}) \leftarrow &\underset{t \in \mathcal{U}, b_{t}^{o}}{\arg\min}\Big\{(1+\mu\|b_{t}^{o}\|_{1})  \mathcal{L}(w_{t} )   \nonumber 
 \\& +\lambda \sum_{s \in \mathcal{U}  -  t}\|w_{s}-\sum_{j=1}^{i-1} B_{\pi(j) s} w_{\pi(j)}-B_{t s} w_{t}\|_{2}^{2}\Big\}. \label{eq:5}
\end{align}
After selecting the task to be learned next and updating $b_{t}^{o}$ by performing alternating optimization on Eq. \eqref{eq:5}, Lee et al.   \cite{lee2016asymmetric} solved  problem Eq. \eqref{eq:3} to greedily minimize Eq. \eqref{eq:2}:
\begin{align}
w_{t} \leftarrow & \underset{w_{t}}{\arg\min}\Big\{(1+\mu\|b_{t}^{o}\|_{1})  \mathcal{L}(w_{t} ) \nonumber  \\&+\lambda\|w_{t}-\sum_{j=1}^{i-1} B_{\pi(j) t} w_{\pi(j)}\|_{2}^{2}\Big\}. \label{eq:3}
\end{align}

\subsection{Loss Modification}

To reasonably distribute the count of transfers between every two tasks and  avoid too many negative transfers, 
we have the following expectations and we modify the loss term in CAMRL based on these  expectations :

     \textbf{[Expectation 1]} Denote $p_{t,i}$ as the performance, usually the average rewards over several evaluation episodes, on training task $i$ by using the network originally for training task $t$. When training task $t$, if we can test the transferability among several tasks and approximately obtain $p_{t,i_1}>p_{t,i_2}>\cdots>p_{t,i_q}$ for some tasks $i_1,i_2,\cdots,i_q$, then we expect that $B_{t,i_1}>B_{t,i_2}>\cdots>B_{t,i_q}$ holds as much as possible, that is, the better evaluation performance on other tasks, the larger number of transfers on those tasks. 
     To meet the above expectation, we expect
     $\sum_{j=1}^{q}(j-y_{i_j})^2$ to be as small as possible, where $j$ is the ranking of $p_{t,i_j}$ among
    $p_{t,i_1}>p_{t,i_2}>\cdots>p_{t,i_q}$
    and
    $y_{i_j}$ is the ranking of $B_{t,i_j}$ among $B_{t,i_1},B_{t,i_2},\cdots,B_{t,i_q}$. 
    
    In the normal case, $y_{i_j}$ is non-differentiable on $B_{t,i_1},B_{t,i_2},\cdots,B_{t,i_q}$. Therefore, we construct a novel differentiable ranking function $y^{\prime}_{i_j}=q+1-\sum_{s=1}^{q}\{0.5*\tanh [d(B_{t,i_j}-B_{t,i_s})]+0.5 \}$ to replace $y_{i_j}$ and the ranking loss thus becomes $\sum_{j=1}^{q}(j-y^{\prime}_{i_j})^2$.

   Here  note that
    the idea of our differentiable ranking function is inspired by  \cite{DBLP:journals/corr/abs-1904-02345}. Ayman \cite{DBLP:journals/corr/abs-1904-02345} customized an activation function in the form of multiple $\tanh$ functions, which   approximates 
 the step function with equidistant points. Here we modify the intercept representation according to different combinations for the point set so as to obtain our differentiable ranking loss which allows  cut-off  points with unequal distance. 
 With our modification, the ranking loss can  incorporate various  prior knowledge and training factors to avoid negative transfer, besides   the relative training difficulty and  the  performance of mutual evaluation between tasks, as stated below.


\begin{figure}[t] 
\vspace{-0.4cm}
     \centering
\includegraphics[width=0.7\linewidth]{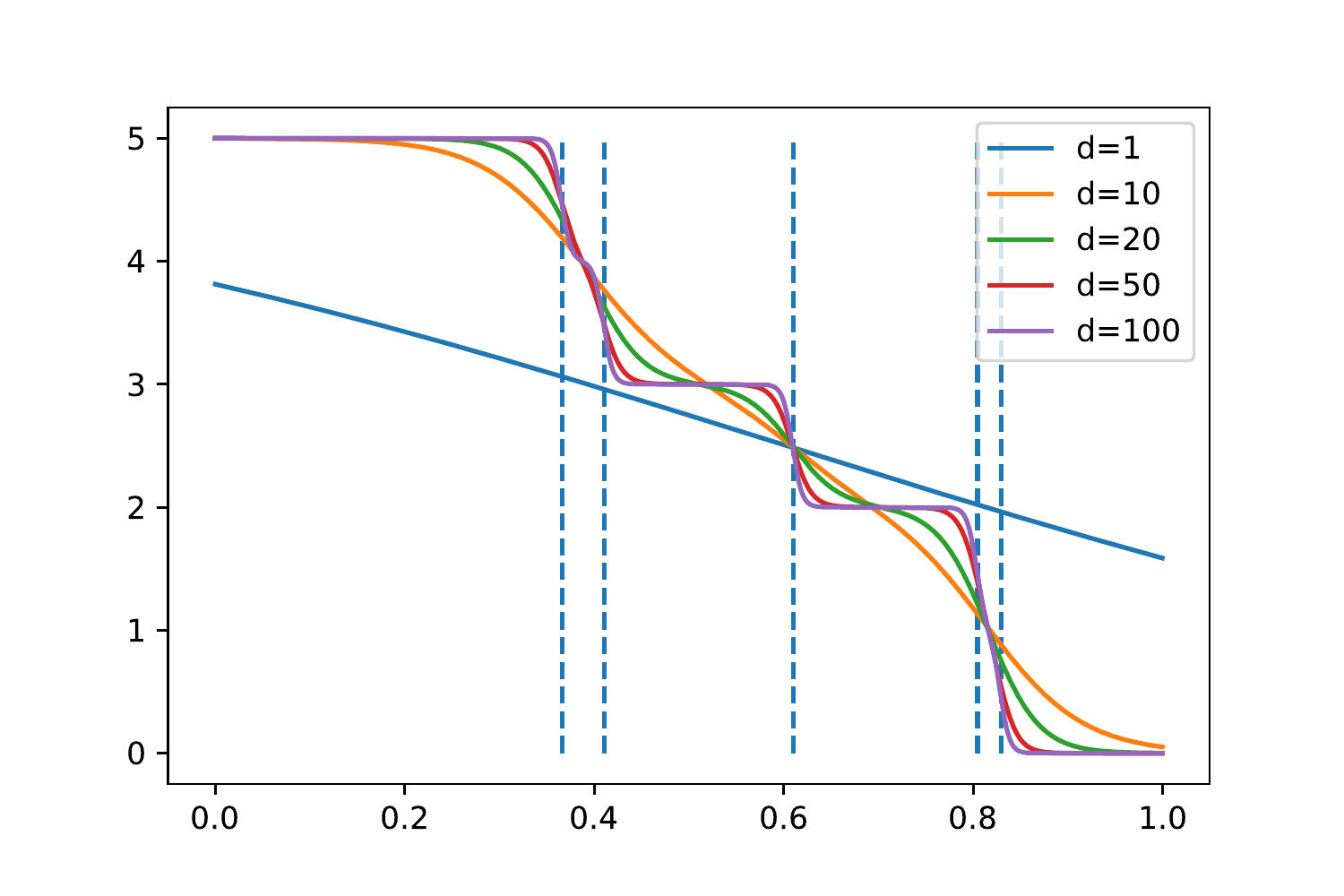}
\caption{Ranking functions with various $d$.  The points displayed by the dashed lines refer to our randomly generated $\{p_{t,i_j}\}_{j\in [q]}$.}
    \label{fig:tanh}
\vspace{-0.3cm}
\end{figure} 

Figure \ref{fig:tanh} depicts the shapes of our adapted composite $\tanh$ function under different  $d$. At the end of every epoch with length $K$, we can randomly select some tasks, measure  the performance of the network which is originally trained for task $i$ on task $j$, and update  $p_{i,j}$.
   
  \textbf{[Expectation 2] } If task $i$ is easier to train, that is, with a smaller $\mathcal{L}(w_i)$, then we expect a smaller number of transfers from task $t$ to task $i$. This is because  if a task is easy-to-train, i.e., with a faster training progress than other tasks, then the number of transfer to that task from other hard-to-train (i.e.,  with slower training progress) tasks should intuitively be smaller.
  To achieve this, we define $l_t^o$ as $(\mathcal{L}(w_1), \ldots, \mathcal{L}(w_{t-1}), \mathcal{L}(w_{t+1}), \ldots, \mathcal{L}(w_T))$. Then add the $-(b_t^o)^\top l_t^o$ term 
 and the corresponding differentiable ranking loss $\sum_{j\in [T]}(rank^{(1)}_j-y{\prime\prime}_{j})^2$ to Eq. \eqref{eq:3}, where $rank^{(1)}_j$ is the ranking of $\mathcal{L}(w_j)$ among
    $\mathcal{L}(w_1),\mathcal{L}(w_2),\cdots,\mathcal{L}(w_T)$
    and    $y{\prime\prime}_{j}$ is the ranking of $B_{t,j}$ among $B_{t,1},B_{t,2},\cdots,B_{t,T}$. 
    
 Note that
 the reason to add the $-(b_t^o)^\top l_t^o$ is that to minimize $-(b_t^o)^\top l_t^o$, we need to maximize $(b_t^o)^\top l_t^o$. The maximum  y $(b_t^o)^\top l_t^o$ corresponds to the scenario where larger $B_{t,i}$ is multiplied with larger $\mathcal{L}(w_i)$ and smaller  $B_{t,i}$ is multiplied with smaller $\mathcal{L}(w_i)$, which perfectly matches our second expectation. 
 
\textbf{[Expectation 3]} If  task $i$ is more similar to task $t$, then we expect a larger number of transfers from task $t$ to task $i$. Denote the similarity between task $i$ and task $t$ as $s_{i,t}$ and $rank^{(2)}_i$ as  the ranking of $s_{i,t}$ among
    $(s_{1,t},s_{2,t},\cdots,s_{T,t})$. Then in summary, the  problem in Eq. \eqref{eq:5} now becomes:
\begin{align}
&(t, b_{t}^{o}) \!\leftarrow \!\underset{t \in \mathcal{U}, b_{t}^{o}}{\arg\min}\Big\{\lambda_0[(1\!+\!\!\mu_1\|b_{t}^{o}\|_{1}) \mathcal{L}(w_{t} ) \!-\!\mu_2 (b_t^o)^\top l_t^o]  \!\nonumber \\&+\!\lambda_1 \!\sum_{s \in \mathcal{U}  -  t}\Big\|w_{s}\!-\!\!\sum_{j=1}^{i-1} B_{\pi(j) s} w_{\pi(j)}\!-\!B_{t s} w_{t}\Big\|_{2}^{2}  +\lambda_2 \sum_{j\in [q]}(j-y{\prime}_{i_j})^2 \nonumber\\&+\lambda_3 \sum_{j\in [T]}(rank^{(1)}_j-y{\prime\prime}_{j})^2  +\lambda_4 \sum_{j\in [T]}(rank^{(2)}_j-y{\prime\prime}_{j})^2
\Big\}. \label{eq:15}
\end{align}

In order to  optimize the objective in Eq. \eqref{eq:15}, we first fix $b_{t}^{o}$ and select task $t$ with the minimum objective in Eq. \eqref{eq:15}. Then fix $t$ and apply the Frank-Wolfe algorithm to optimize $b_{t}^{o}$  with a convergence guarantee. Finally, fix $b_{t}^{o}$ and train task $t$  with the policy loss  Eq. \eqref{eq:7} for $K$ episodes. 

\begin{remark}
    \textbf{Similarity measures for expectation 3.}
    Regarding the expectation 3 for our customized ranking loss function,
 we take the  following three similarity measures, respectively:
 a) the cosine similarity between the critic networks; b) the cosine similarity between the policy networks;  c) the negative embedding distance in the state space, which will be explained in the Discussion Section. Among them, the first measure has the best performance in  initial trials and hence we utilize this measure in all our experiments.
As for the negative embedding distance between task $i$ and task $j$,
denote the state space as $\mathcal{S}$ and the embedding distance between task $i$ and task $j$ is calculated as follows:
(i)
Sample $100$ states, $\{s_1,s_2,\cdots,s_{100}\}$, from $\mathcal{S}$  uniformly at random.
(ii) Denote the embeddings of  $\{s_1,s_2,\dots,s_{100}\}$ calculated by the embedding layers of task $i$ and task $j$ as $\{e_{i,1},e_{i,2},\cdots,e_{i,100}\}$ and $\{e_{j,1},e_{j,2},\cdots,e_{j,100}\}$, respectively.
(iii) Then, $-\sqrt{\frac{1}{100}\sum_{m=1}^{100}||e_{i,m}-e_{j,m}||_2^2}$ is  just the negative embedding distance between task $i$ and task $j$.
\end{remark}

\subsection{Loss Optimization}

\textbf{Optimization of $b_{t}^{o}$.}\ 
To make the loss  differentiable and ensure the sparsity of the transfer matrix $B$, we 
replace $\mu_1\|b_{t}^{o}\|_{1}$ in Eq. \eqref{eq:15} with $\mu_1 \sum_{j\in [T] -  \{t\}} B_{tj}$ and add the $b_{t}^{o}\geq 0, \|b_{t}^{o}\|_{1}\leq radius<\frac{1}{2}$ constraints to Eq. \eqref{eq:15}, where $radius$ is the upper bound of $\|b_{t}^{o}\|_{1} (t\in [T])$ that we set in experiments. Define
  $f(b_{t}^{o})=\lambda_0[(1+\mu_1 \sum_{j\in [T] -  \{t\}} B_{tj})  \mathcal{L}(w_{t} ) -\mu_2 (b_t^o)^\top l_t^o ] +
\lambda_1 \sum_{s \in \mathcal{U}  -  t}\|w_{s}-\sum_{j=1}^{i-1} B_{\pi(j) s} w_{\pi(j)}-B_{t s} w_{t}\|_{2}^{2}
+\lambda_2 \sum_{j\in [q]}(j-y{\prime}_{i_j})^2+\lambda_3 \sum_{j\in [T]}(rank^{(1)}_j-y{\prime\prime}_{j})^2 +\lambda_4 \sum_{j\in [T]}(rank^{(2)}_j-y{\prime\prime}_{j})^2$, and we can transform the optimization on $b_{t}^{o}$ into the following constrained optimization problem with objective $f(b_{t}^{o})$:
\begin{align}\label{eq:6}
    \underset{b_{t}^{o}\geq 0, \|b_{t}^{o}\|_{1}\leq radius}{\min} f(b_{t}^{o}).
\end{align}
 
In experiments, we apply the following iterative methods
to solve the constrained optimization problem \eqref{eq:6}: 
 vanilla Frank–Wolfe \cite{freund2016new}, momentum Frank–Wolfe \cite{constopt-pytorch}, projected gradient descent (PGD), PGDMadry \cite{madry2017towards},  and the General Iterative Shrinkage and Thresholding algorithm (GIST)  \cite{gong2013general}. Among them, vanilla Frank–Wolfe achieves smaller $f(b_{t}^{o})$  within the same number of iterations  and better convergence performance. Therefore,
we use vanilla Frank–Wolfe to optimize $b_{t}^{o}$ with the convergence rate  at the $m$-th  iteration $\frac{1}{\sqrt{m}}$, according to \cite{Lacoste} and \cite{Nesterov18}. See  Appendix for the proof.
 
 \textbf{Dynamic adjustment of hyper-parameters.}\ 
 We take the optimization of each term in Eq. \eqref{eq:15} as a multi-task learning problem and automatically adjust $\{\lambda_i\}_{i=0}^{4}$ according to each term's standard error over the historical epochs \cite{kendall2018multi}. Note that Kendal et al. \cite{kendall2018multi} derived their multi-task loss function based on maximizing the Gaussian likelihood with homoscedastic
uncertainty on supervised learning tasks, due to the popularity and the simplicity of the  method in \cite{kendall2018multi}. 
Here we adapt this method by refining the denominators in the coefficients of the loss to deal with our setting.   

To be more specific,
similar to \cite{kendall2018multi},   let $\lambda_i=\frac{1}{4\sigma_i^2+\epsilon} (i\in [4])$ and $\lambda_0=\frac{1}{2\sigma_0^2+\epsilon}$, where $\sigma_i$  is the standard error of the $(i+1)$-th term in Eq. \eqref{eq:15}, $\epsilon (=10^{-2})$ is added to avoid the zero value of denominators, and we set $\lambda_0$ to be $\frac{1}{2\sigma_0^2}$ instead of $\frac{1}{4\sigma_0^2}$ to pay more attention to the original training loss. Thanks to this automatic adjustment, our CAMRL eliminates the need for laborious hyper-parameter analysis.

\section{Experiments}

\begin{table*}
\centering
\scriptsize
\caption{Averaged reward for Gym-minigrid tasks (each experiment repeated $5$ times, the average scores with standard errors (in brackets) reported). The best results are bolded. The higher the metric, the better performance of the model.}  \label{tab:minigrid}
\begin{tabular}{lccccccc}
\toprule
 & Single AC    & Single AC+MRCL   & YOLOR-SAC+MRCL   & Distral+MRCL    & Gradient Surgery+MRCL   & CAMRL+MRCL(ours) \\ \midrule
DistShiftEnv             & 0.05             (0.02)           & 0.78                  (0.05)               & 0.09            (0.03)          & 0.07              (0.04)            & 0.08                        (0.02)                     & \textbf{0.95}                   (0.03)              \\
DoorKeyEnv16x16          & 0.00             (0.00)            & 0.08                  (0.03)               & 0.01             (0.01)            & 0.00                (0.00)               & 0.01                         (0.01)                        & \textbf{0.13   }                (0.02)                  \\
DynamicObstaclesEnv16x16 & -0.99           (0.03)            & -0.83                 (0.04)                & -0.88            (0.03)            & -0.98               (0.01)               & -0.97                        (0.02)                        & \textbf{-0.01   }               (0.02)                  \\
EmptyEnv16x16            & 0.10             (0.04)            & 1.10                  (0.04)                 & 0.12             (0.03)            & 0.08                (0.03)               & 0.11                         (0.03)                        & \textbf{1.17 }                  (0.05)                  \\
FetchEnv                 & 0.07             (0.03)            & 0.83                  (0.05)                 & 0.14             (0.03)            & 0.09                (0.02)               & 0.09                         (0.03)                        & \textbf{0.89 }                  (0.04)                  \\
KeyCorridor              & 0.01             (0.01)            & 0.02                  (0.01)                 & 0.09             (0.03)            & 0.02                (0.01)               & 0.08                         (0.02)                        & \textbf{0.51 }                  (0.02)                  \\
LavaCrossingS9N3Env      & 0.02             (0.01)            & \textbf{0.13  }                (0.02)                 & 0.02             (0.01)            & 0.01                (0.01)               & 0.02                         (0.01)                        & 0.04                   (0.01)                  \\
LavaGapS7Env             & 0.03             (0.02)            & 0.53                  (0.04)                 & 0.21             (0.02)            & 0.03                (0.01)               & 0.05                         (0.02)                        & \textbf{0.87}                   (0.02)                  \\
MemoryS17Random          & 0.02             (0.01)            & \textbf{0.61 }                 (0.04)                 & 0.13             (0.02)            & 0.04                (0.02)               & 0.20                         (0.04)                        & 0.20                   (0.03)                    \\   
\bottomrule
\end{tabular}
\end{table*}

\begin{table*}
\centering
\scriptsize
\caption{Averaged reward for MT10 tasks.  The higher the metric, the better performance of the model.}   \label{tab:mt10}

\begin{tabular}{lccccccc}
\toprule
& Single AC & Single SAC & YOLOR-SAC & Distral  & Gradient Surgery & Soft Module & CAMRL(ours) \\
\midrule
Pick and place     & 24277.65 (2349.23) & 38526.19 (1930.41) & 1744.85 (98.42) & 12791.42 (938.4) & 4347.38 (294.3) & 324.91   (40.91)  & \textbf{47235.31 }(1043.29) \\
Pushing            & -70.93   (10.26)   & 5.39    (4.93)    & -10.93  (8.54)  & -63.57   (4.82)  & -127.42 (4.25)  & -53.84   (4.82)   &\textbf{ 213.27}   (10.51)   \\
Reaching           & -109.39  (9.74)    & -49.73   (5.91)    & -53.31  (5.04)  & -123.61  (6.25)  & -117.74 (4.02)  & \textbf{-38.36 }  (5.39)   & -50.36   (2.55)    \\
Door opening       & -50.31   (8.22)    & 17.72    (7.14)    & 10.48   (6.41)  & -55.41   (4.92)  & -38.73  (5.49)  & 13.86    (6.03)   & \textbf{143.74}   (10.48)   \\
Button press       & -28.94   (3.04)    & \textbf{2539.78}  (214.49)  & 409.21  (4.38)  & -15.87   (3.97)  & -63.25  (7.93)  & -9.41    (3.15)   & 1395.77  (82.91)   \\
Peg insertion side & -78.84   (5.21)    & 25.64    (6.92)    & 20.31   (4.27)  & -124.82  (7.92)  & -79.74  (5.37)  & -73.74   (6.93)   & \textbf{52.91}    (3.81)    \\
Window opening     & 13.82    (2.48)    & 1937.28  (102.41)  & 105.38  (18.49) & 11.98    (3.51)  & 8.93    (3.85)  & 9.21     (4.14)   & \textbf{15683.29} (1702.03) \\
Window closing     & 9.74    (2.31)    & 8.96      (2.18)    & 15.87   (4.66)  & 10.71    (2.04)  & 12.82   (3.02)  & \textbf{37726.31} (2049.4) & 23.64    (3.25)    \\
Drawer opening     & -19.83   (3.59)    & 385.72   (39.03)   & 582.13  (48.93) & -17.82   (6.15)  & -9.32   (3.61)  & 573.82   (30.98)  & \textbf{1329.39}  (28.94)   \\
Drawer closing     & \textbf{1532.98}  (234.26)  & 310.83   (14.09)   & 193.84  (16.02) & -40.86   (12.06) & -39.94  (4.83)  & 611.83   (38.94)  & 622.74   (10.93)  \\
\bottomrule
\end{tabular} 
\end{table*}

\begin{table*}
\centering
\scriptsize
\caption{Averaged reward for MT50 tasks.  The higher the metric, the better performance of the model.} \label{tab:mt50_whole2}

\begin{tabular}{lccccccc}
\toprule
& Single AC & Single SAC & YOLOR-SAC  & Distral & Gradient Surgery & Soft Module & CAMRL(ours) \\ \midrule
Turn on faucet          & \textbf{49832.92} (3903.31) & 4793.37 (203.94)  & 1495.39 (103.47) & 7193.81 (102.78) & 2893.73 (49.01)  & 2281.66 (56.93)  & 7644.89  (61.81)  \\
Sweep                   & -83.03   (20.93)   & -39.84  (7.93)    & -94.11  (8.33)   & -112.93 (13.95)  & -83.56  (15.03)  & -18.38  (5.03)   &\textbf{ -10.31   }(3.19)   \\
Stack                   & -110.28  (19.04)   & -110.38 (8.94)    & -87.39  (7.37)   & -152.85 (14.91)  & -103.95 (7.53)   & -46.78  (5.82)   & \textbf{-42.32 }  (4.01)  \\
Unstack                 & -47.39   (8.05)    & -47.17  (9.02)    & -56.31 (6.49)   & -74.93  (6.93)   & -62.38  (10.05)  & -54.21  (6.18)   & \textbf{-31.26 }  (5.09)   \\
Turn off faucet         & -50.89   (4.95)    & -8.38   (3.75)    & 83.92   (4.28)   & -12.93  (5.03)   & -2.31   (4.62)   & 24.49   (6.83)   & \textbf{1253.73 } (21.15)  \\
Push back               & -52.04   (10.03)   & -38.93  (6.08)    & -39.02  (4.39)   & -122.84 (4.95)   & -98.53  (7.93)   &\textbf{ 27.36  } (12.05)  & -27.74   (3.16)   \\
Pull lever              & -82.94   (10.92)   & -63.41  (8.38)    & -68.41  (8.02)   & -92.38  (7.31)   & -113.26 (13.85)  & -69.45  (9.96)   & \textbf{-31.84 }  (7.06)   \\
Turn dial               & -68.31   (9.83)    & -50.39  (7.17)    & -50.37  (8.36)   & -130.36 (11.03)  & -58.37  (7.84)   & -29.32  (4.01)   &\textbf{ -27.69   }(7.68)   \\
Push with stick         & 703.34   (38.98)   & -5.81   (1.83)    & 102.49  (14.07)  & -15.83  (2.97)   & -14.92  (4.04)   & -32.89 (5.81)  & \textbf{1376.83}  (38.44)  \\
Get coffee              & -74.95   (10.95)   & 72.18   (10.84)   & -42.46  (9.33)   & -142.18 (7.89)   & -40.35  (4.82)   & \textbf{-12.96 } (7.37)   & -52.93   (4.91)   \\
Pull handle side        & -51.59   (10.02)   & -17.29  (8.05)    & -59.02  (5.12)   & \textbf{4938.29} (219.77) & -62.83  (13.05)  & -58.72  (14.73)  & -48.93   (5.89)   \\
Basketball              & 3295.06  (203.9)   & 2507.48 (114.85)  & 1830.94 (88.26)  & -163.21 (30.46)  & 4817.61 (248.49) & 2194.36 (305.19) & \textbf{16394.92} (319.85) \\
Pull with stick         & -123.59  (8.21)    & -40.09  (6.73)    & -38.05  (5.62)   & -113.48 (7.99)   & -118.37 (8.38)   & -38.91  (7.74)  & \textbf{-28.92}   (5.13)   \\
Sweep into hole         & -82.83   (8.48)    & -38.34  (7.92)    & -10.99  (4.78)   & 10.76   (8.19)   & -38.96  (11.01)  & -28.53  (6.93)   & \textbf{16.74}    (5.96)   \\
Disassemble nut         & 39.09    (7.32)    & \textbf{3947.53} (393.07) & 102.58  (7.49)   & 20.48   (5.68)   & -3.15   (4.82)   & 2160.02 (117.83) & 1873.93  (79.22)  \\
Place onto shell        & \textbf{11093.45} (302.38)  & 2.94    (4.02)    & 183.73  (19.41)  & -50.31  (5.24)   & 6973.13 (294.05) & 11.97   (5.93)   & 274.01   (23.91)  \\
Push mug                & -30.19   (5.31)    & 40.35   (7.08)   & -102.29 (9.28)   & -227.48 (5.39)   & -93.28  (8.29)   & 119.93  (16.34)  & \textbf{638.91 }  (11.34)  \\
Press handle side       & -127.49  (8.91)    & -61.31  (5.05)    & 10.21   (6.77)   & -173.83 (6.03)   & -117.31 (10.93)  & -61.28  (4.91)   & \textbf{24.18 }   (5.03)   \\
Hammer                  & -74.82   (5.04)    & -73.29  (4.91)    & -79.63  (8.29)   & -147.21 (5.83)   & -91.22  (6.85)   & -71.98  (4.05)   & \textbf{-60.93 }  (3.92)   \\
Slide plate             & -78.25   (7.93)    & -39.14  (7.03)    & -0.92   (5.28)   & -122.84 (10.04)  & -107.31 (8.86)   & -33.71  (5.19)   & \textbf{425.64}   (10.62)  \\
Slide plate side        & -81.43   (6.37)    & \textbf{-10.28}  (3.09)    & -72.28  (6.99)   & -103.95 (7.01)   & -92.18  (10.51)  & -25.62  (4.97)   & -23.54   (3.09)   \\
Press button wall       & -104.18  (6.55)    & -56.49  (4.72)    & -67.27  (8.19)  & -110.38 (4.96)   & -89.17  (7.31)   & \textbf{64.83}   (7.37)   & -27.73   (4.16)   \\
Press handle            & -107.74  (7.18)    & \textbf{-39.26}  (4.27)    & -68.81  (11.23)  & -106.91 (4.08)   & -65.28  (6.83)   & -48.36  (5.15)   & -49.18   (4.92)   \\
Pull handle             & -82.94   (5.03)    & 34.58   (4.86)    & -82.17  (7.28)   & -134.02 (5.82)   & -77.29  (5.93)   & -22.16  (4.17)   & \textbf{46.84  }  (5.86)   \\
Soccer                  & -72.38   (8.93)    & 429.94  (20.28)   & 102.36  (9.66)   & -108.83 (5.37)   & -9.56   (3.85)   & 9.68    (3.01)   & 4\textbf{85.03   } (20.75)  \\
Retrieve plate side     & -87.36   (16.83)   & -3.91   (4.92)    & -1.38   (3.94)   & -105.31 (6.77)   & -78.93  (6.42)   & -96.43  (8.93)   & \textbf{127.94}   (12.84)  \\
Retrieve plate          & -116.93  (10.35)   & -1.63   (5.03)    & -45.08  (5.42)   & -123.98 (7.94)   & -62.18  (7.39)   & \textbf{24.91 }  (5.06)   & -42.65   (6.27)   \\
Close drawer            & -58.81   (5.61)    & -0.98   (4.75)    & 29.03   (6.12)   & -114.05 (7.13)   & -94.77  (5.74)   & 31.58   (4.97)   & \textbf{428.93}   (29.04)  \\
Press button top        & -107.47  (7.09)    & -42.38  (5.38)    & -30.81  (6.17)   & -118.37 (4.93)   & -106.31 (5.28)   & -37.23  (7.08)   & \textbf{-29.84 }  (4.17)   \\
Reach                   & -101.75  (8.94)    & -37.01  (4.93)    & -48.92  (7.55)   & -121.94 (5.88)   & -102.67 (6.24)   & -38.77  (4.03)   & \textbf{-23.47 }  (4.22)   \\
Press button top w/wall & -109.28  (5.93)    & -45.02  (6.02)    & -60.39  (7.82)   & -162.38 (6.85)   & -94.29  (4.91)   & \textbf{-37.39}  (5.42)   & -41.84   (4.38)   \\
Reach with wall         & 0.27     (0.13)    & 2.97    (1.02)    & 30.44   (5.18)   & 20.36   (4.39)   & 3.68    (1.81)   & 23.61   (4.95)   & \textbf{897.42}   (39.57)  \\
Insert peg side         & 10.84    (4.05)    & 280.17  (25.91)   & 9.39    (3.17)   & -12.89  (4.03)   & 237.51  (18.32)  & \textbf{832.85 } (68.03)  & 12.93    (3.94)   \\
Push                    & -33.48   (5.06)    & -23.65  (4.09)    & 7.49    (5.04)   & -64.58  (6.28)   & -23.87  (4.02)   & -72.56  (8.92)   & \textbf{573.94 }  (28.05)  \\
Push with wall          & -75.82   (8.96)    & -42.71  (4.62)    & -49.07  (7.33)   & -105.37 (6.94)   & -76.32  (5.35)   & -48.38  (6.52)   & \textbf{23.81}    (4.13)   \\
Pick \& place w/wall    & -53.14   (5.53)    & -43.95  (4.29)    & -55.39  (6.31)   & -89.96  (7.32)   & -57.94  (8.56)   & -47.32  (7.44)   & \textbf{-33.27}   (5.24)   \\
Press button            & 7795.45  (602.42)  & \textbf{7882.14} (595.06)  & 109.27  (13.49)  & -10.46  (4.03)   & 28.35   (5.91)   & 9.86    (5.68)   & 847.93   (37.26)  \\
Pick \& place           & -92.49   (6.75)    & -40.29  (5.62)    & -40.88  (7.32)   & -118.91 (8.52)   & -84.74  (6.31)   & -44.58  (4.74)   & \textbf{-23.15 }  (4.51)   \\
Pull mug                & -37.03   (5.67)    & -18.93  (4.67)    & -50.36  (6.58)   & -104.29 (7.43)   & -53.27  (5.46)   & -34.01  (4.09)   & \textbf{-3.03   } (3.56)   \\
Unplug peg              & -82.37   (8.53)    & -73.28  (6.75)    & \textbf{-49.13}  (8.33)   & -116.72 (8.52)   & -117.64 (5.99)   & -63.28  (4.76)   & -52.94   (4.73)   \\
Close window            & -58.32   (5.93)    & -32.19  (4.14)    & -40.92  (5.17)   & -113.96 (6.46)   & -54.73  (5.35)   & -28.94  (4.78)   & \textbf{-21.19 }  (4.22)   \\
Open window             & -143.54  (6.67)    & -39.65  (5.46)    & -57.41  (6.35)   & -78.67  (6.74)   & -129.86 (7.46)   & -33.42  (5.91)   & \textbf{-27.36}   (5.57)   \\
Open door               & -83.84   (5.35)    & -64.36  (6.77)    & -76.57  (10.27)  & -153.26 (5.04)   & -98.69  (6.53)   & -72.91  (5.86)   & \textbf{-62.27}   (5.02)   \\
Close door              & -23.95   (6.43)    & -35.63  (7.93)    & -34.93  (6.74)   & -87.69  (5.64)   & -88.72  (6.08)   & -27.83  (5.83)   & \textbf{-23.01}   (4.98)   \\
Open drawer             & -62.17   (5.59)    & -28.95  (6.43)    & -71.49  (8.37)   & -91.24  (6.61)   & -67.62  (6.72)   & \textbf{-25.98}  (5.79)   & -60.03   (5.63)   \\
Open box                & -92.43   (5.89)    & -105.74 (6.75)    & -41.05  (6.07)   & -183.28 (7.62)   & -171.39 (6.35)   & -67.84  (5.82)   & \textbf{-27.09 }  (5.31)   \\
Close box               & -99.31   (7.86)    & \textbf{-26.93}  (6.74)    & -66.38  (7.31)   & -163.62 (9.63)   & -133.86 (6.42)   &\textbf{ -28.21}  (7.03)   & -47.15   (6.46)   \\
Lock door               & -107.16  (10.55)   & -48.69  (9.93)    & -53.19  (4.19)   & -73.25  (5.61)   & -86.32  (5.69)   & -47.24  (6.45)   & \textbf{-45.39}   (4.13)   \\
Unlock door             & -38.43   (4.99)    & -44.31  (5.42)    & -41.48  (4.28)   & -64.59  (5.67)   & -57.24  (7.53)   & -46.29  (5.96)   & \textbf{-33.06}   (6.62)   \\
Pick bin                & -82.96   (5.64)    & -28.93  (4.84)    & -45.38  (5.46)   & -107.51 (6.32)   & -39.18  (4.99)   & \textbf{-16.27}  (5.47)   & -42.94   (5.14)  
\\ \bottomrule
\end{tabular}

\end{table*}

\begin{table*}
\centering
\scriptsize
\caption{Averaged reward for Atari tasks.  The higher the metric, the better performance of the model.} \label{tab:atari}

\begin{tabular}{lccccccc} \toprule
 & Single AC & Single SAC & YOLOR-SAC   & Distral & Gradient Surgery & Soft Module & CAMRL(ours) \\ \midrule
YarsRevenge    & 1208.71 (50.96) & 1293.97 (56.52) & 982.31 (42.38) & 1219.05 (49.64) & 1108.92 (52.69) & 694.21 (38.01) & \textbf{1637.94} (42.83) \\
Jamesbond      & 7.92    (2.91)  & 8.12    (2.86)  & 7.37   (3.02)  & 7.53    (3.06)  & 7.51    (2.17)  & 7.58   (3.42)  & \textbf{8.23}    (2.64)  \\
FishingDerby   & \textbf{-10.18  } (4.03)  & -11.26  (4.57)  & -12.33 (4.27)  & -11.49  (3.98)  & -10.38  (3.63)  & -11.56 (4.02)  & -11.41           (3.71)  \\
Venture        & 0.02    (0.01)  & 0.07    (0.02)  & 0.02   (0.01)  & 0.04    (0.01)  & 0.01    (0.01)  & 0.09  (0.03)  & \textbf{0.22}    (0.03)  \\
DoubleDunk     & -1.39   (0.94)  & -1.34   (1.02)  & -1.36  (0.83)  & -0.96   (0.46)  & -1.44   (0.58)  & -1.42  (0.46)  & \textbf{-0.89}   (0.31)  \\
Kangaroo       & 6.95    (2.05)  & 7.21    (3.47)  & 4.38   (2.12)  & 6.24    (2.74)  & 5.21    (2.95)  & \textbf{9.82}   (3.84)  & 5.74             (2.08)  \\
IceHockey      & -0.46   (0.28)  & -0.45   (0.31)  & -0.52  (0.17)  & -0.51   (0.24)  & \textbf{-0.32}   (0.26)  & -0.53  (0.31)  & -0.49            (0.27)  \\
ChopperCommand & 193.91  (19.48) & 212.54  (24.95) & 217.38 (17.39) & 232.54  (22.81) & 201.96  (24.04) & 242.91 (27.42) & \textbf{351.68}  (29.05) \\
Krull          & 12.59   (3.51)  & 13.24   (4.09)  & 8.28   (3.27)  & 8.38    (3.44)  & \textbf{37.82}   (8.43)  & 10.92  (3.49)  & 9.75             (3.18)  \\
Robotank       & 0.37    (0.06)  & 0.37    (0.05)  & 0.36   (0.04)  & 0.36    (0.04)  & 0.35    (0.05)  & 0.36   (0.05)  & \textbf{0.38}    (0.04) 
\\ \bottomrule \\
\end{tabular}
\end{table*}

\begin{table*}
 \centering
 \scriptsize
\caption{Averaged success rate for Ravens tasks.  The higher the metric, the better performance of the model.} \label{tab:ravens}
\begin{tabular}{lccccccc} \toprule
 & Single AC & Single SAC & YOLOR-SAC   & Distral & Gradient Surgery & Soft Module & CAMRL (ours) \\ \midrule
Block-insertion     & 80.92 (3.94) & 87.95 (1.63) & 88.31 (2.59) & 89.91 (3.47) & 91.82 (3.41) & 89.95 (6.55) & \textbf{94.26} (5.37) \\
Place-red-in-green  & 84.23 (5.05) & 90.32 (6.15) & 89.94 (2.08) & 91.83 (2.16) & 89.18 (3.02) & 90.14 (2.98) & \textbf{93.44} (6.57) \\
Towers-of-hanoi     & 86.85 (3.93) & 88.47 (2.77) & 87.61 (2.76) & 88.32 (3.38) & 92.37 (2.41) & 89.73 (3.34) & \textbf{95.12} (4.64) \\
Align-box-corner    & 84.39 (4.49) & 86.49 (2.74) & 89.01 (2.46) & 87.65 (2.84) & 90.13 (2.55) & 91.27 (2.83) & \textbf{93.43} (1.35) \\
Stack-block-pyramid & \textbf{79.56 }(2.55) & 78.52 (4.43) & 77.03 (1.98) & 77.43 (4.01) & 78.91 (3.82) & 78.24 (4.28) & 77.15          (2.41) \\
Palletizing-boxes   & 87.25 (2.29) & 90.48 (1.98) & 91.38 (2.02) & 91.78 (2.44) & 91.64 (2.27) & \textbf{94.49} (2.53) & 94.32          (2.63) \\
Assembling-kits     & 82.26 (6.31) & 85.92 (3.31) & 88.48 (2.76) & 87.02 (3.89) & 88.03 (3.47) & 92.91 (2.31) & \textbf{93.95} (4.36) \\
Packing-boxes       & 71.55 (2.69) & 75.42 (1.97) & 78.21 (3.79) & 77.94 (4.02) & 77.38 (3.08) & 79.21 (3.57) & \textbf{79.38 }         (3.01) \\
Manipulating-rope   & 83.04 (5.42) & 87.06 (5.39) & 87.28 (3.47) & 87.13 (2.94) & 84.26 (4.46) & 84.02 (3.59) & \textbf{89.27} (2.14) \\
Sweeping-piles      & 85.61 (2.04) & 88.93 (3.63) & 89.14 (2.71) & 90.15 (2.36) & 90.69 (2.35) & 90.84 (2.31) & \textbf{93.64} (3.21)
 \\ \bottomrule \\
\end{tabular}
\end{table*}

\begin{table*}
\centering
\scriptsize
\caption{Averaged success rate for RLBench tasks.  The higher the metric, the better performance of the model.} \label{tab:rlbench}

\begin{tabular}{lccccccc}
\toprule     
& Single AC & Single SAC & YOLOR-SAC & Distral  & Gradient Surgery & Soft Module & CAMRL(ours) \\
\midrule
Reach Target                & 96.49     (0.83) & 96.41      (0.76) & 96.57     (0.63) & 95.92   (0.67) & 95.08            (0.91) & 96.31       (0.42) & \textbf{99.87    }   (0.05) \\
Push Button                 & 86.49     (2.47) & 87.41      (2.76) & 88.57     (3.04) & 87.38   (2.54) & 84.48            (3.42) & 89.48       (2.35) & \textbf{97.24  }     (0.85) \\
Pick And Lift               & 86.36     (2.02) & 86.49      (2.76) & 87.03     (1.84) & 85.47   (2.63) & 85.31            (2.42) & 87.21       (2.43) & \textbf{91.58}       (1.54) \\
Pick Up Cup                 & 81.47     (3.04) & 81.84      (2.71) & 82.59     (2.46) & 83.02   (2.47) & 83.48            (1.59) & 84.03       (2.41) & \textbf{86.48}       (2.57) \\
Put Knife on Chopping Board & 45.49     (3.19) & 44.54      (3.58) & 46.61     (3.59) & 46.02   (4.47) & 45.58            (3.17) & 46.94       (3.41) & \textbf{50.91 }      (4.48) \\
Take Money Out Safe         & 61.48     (3.18) & 60.45     (3.82) & 61.41     (3.46) & 59.48   (3.77) & 57.57            (3.28) & 59.43       (3.91) & \textbf{66.84  }     (3.56) \\
Put Money In Safe           & 75.35     (3.17) & 74.35      (3.46) & 72.59     (5.57 & 75.48   (3.51) & 73.18            (3.05) & 74.13       (3.06) & \textbf{80.38   }    (2.51) \\
Pick Up Umbrella            & 72.32     (3.31) & 73.58      (3.87) & 75.03     (2.86) & 73.93   (3.01) & 76.57            (3.25) & 76.43       (2.76) & \textbf{81.49 }      (2.95) \\
Stack Wine                  & 17.49     (1.38) & 17.52      (1.75) & 20.53     (2.52) & 19.57   (3.22) & 18.49            (2.63) & 17.56       (2.55) & \textbf{24.59  }     (2.04) \\
Slide Block To Target       & 71.58     (2.59) & 73.69      (3.02) & 75.57     (2.99) & 73.68   (3.57) & 71.01            (3.55) & \textbf{80.17 }      (3.52) & 79.47       (2.57) \\
\bottomrule 
\end{tabular}
\end{table*}

 \begin{figure*}
     \centering
\includegraphics[width=0.24\linewidth]{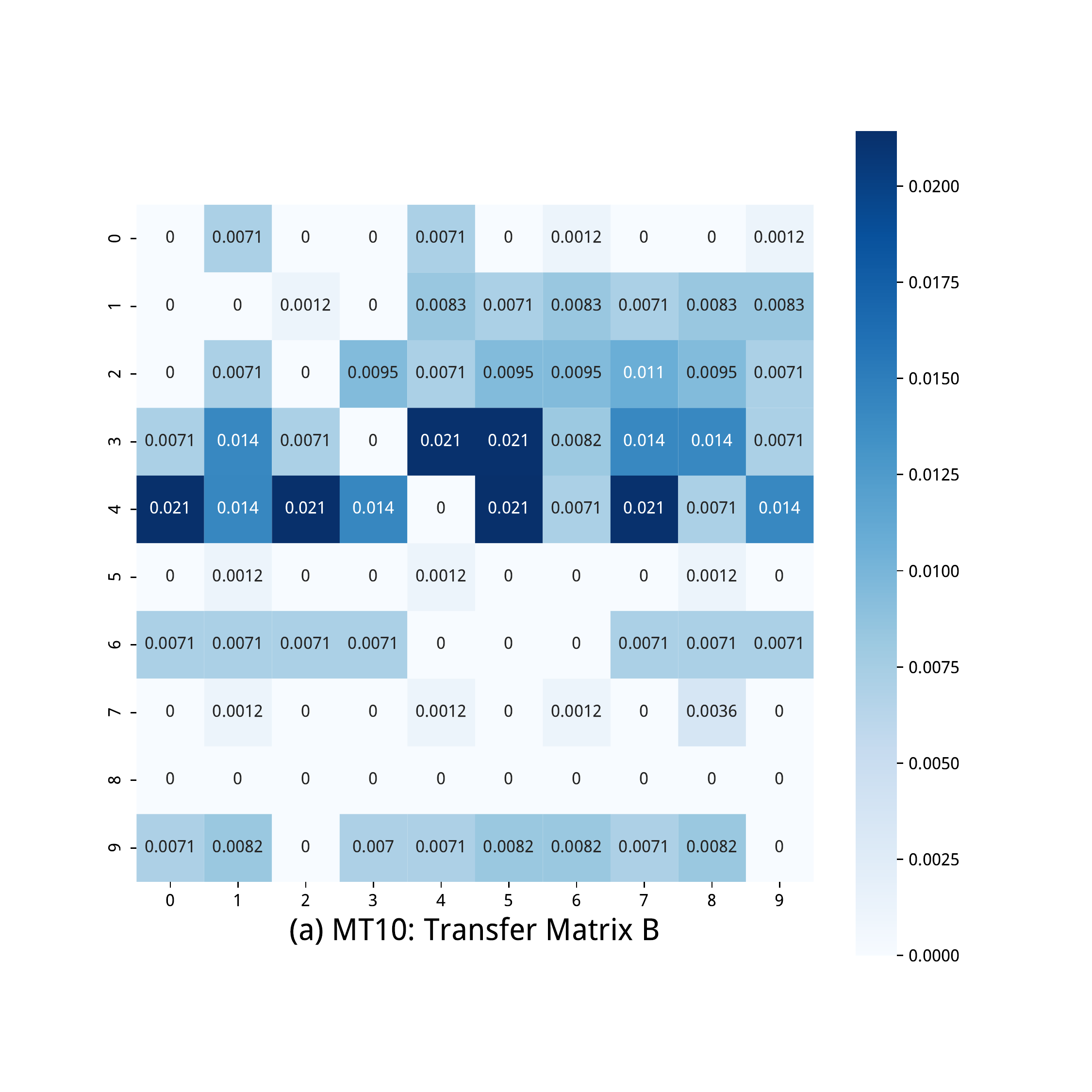} 
\includegraphics[width=0.24\linewidth]{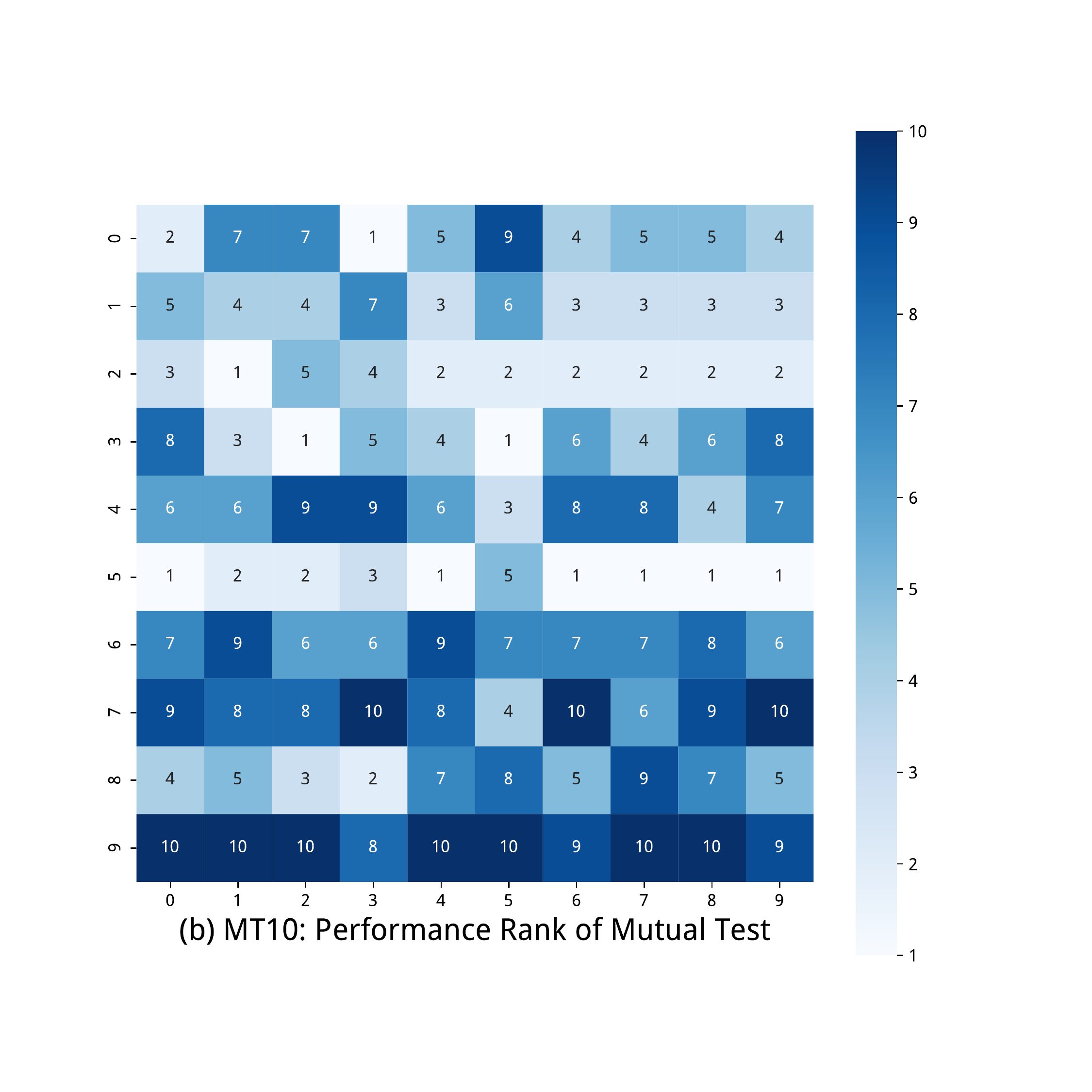} 
\includegraphics[width=0.24\linewidth]{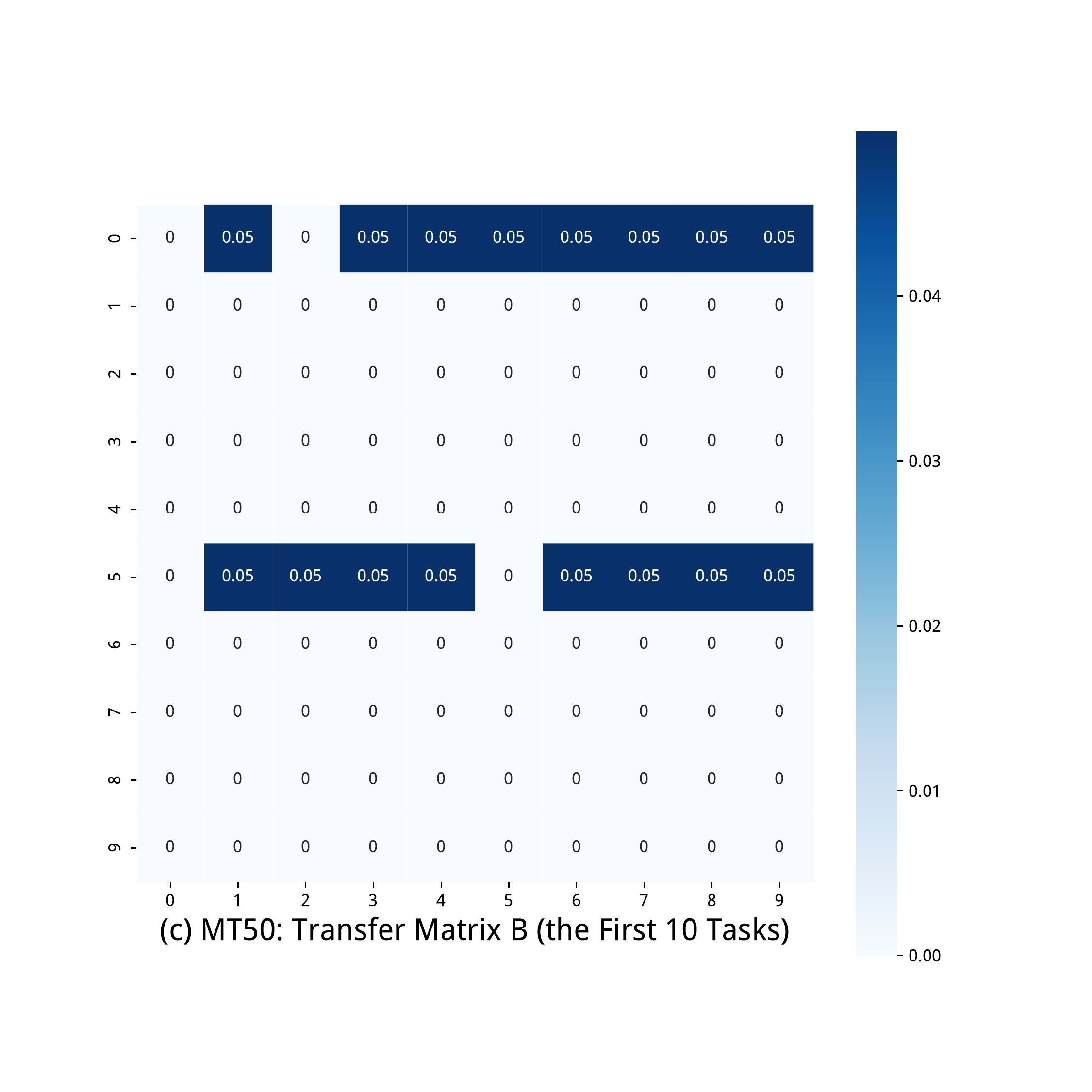} 
\includegraphics[width=0.24\linewidth]{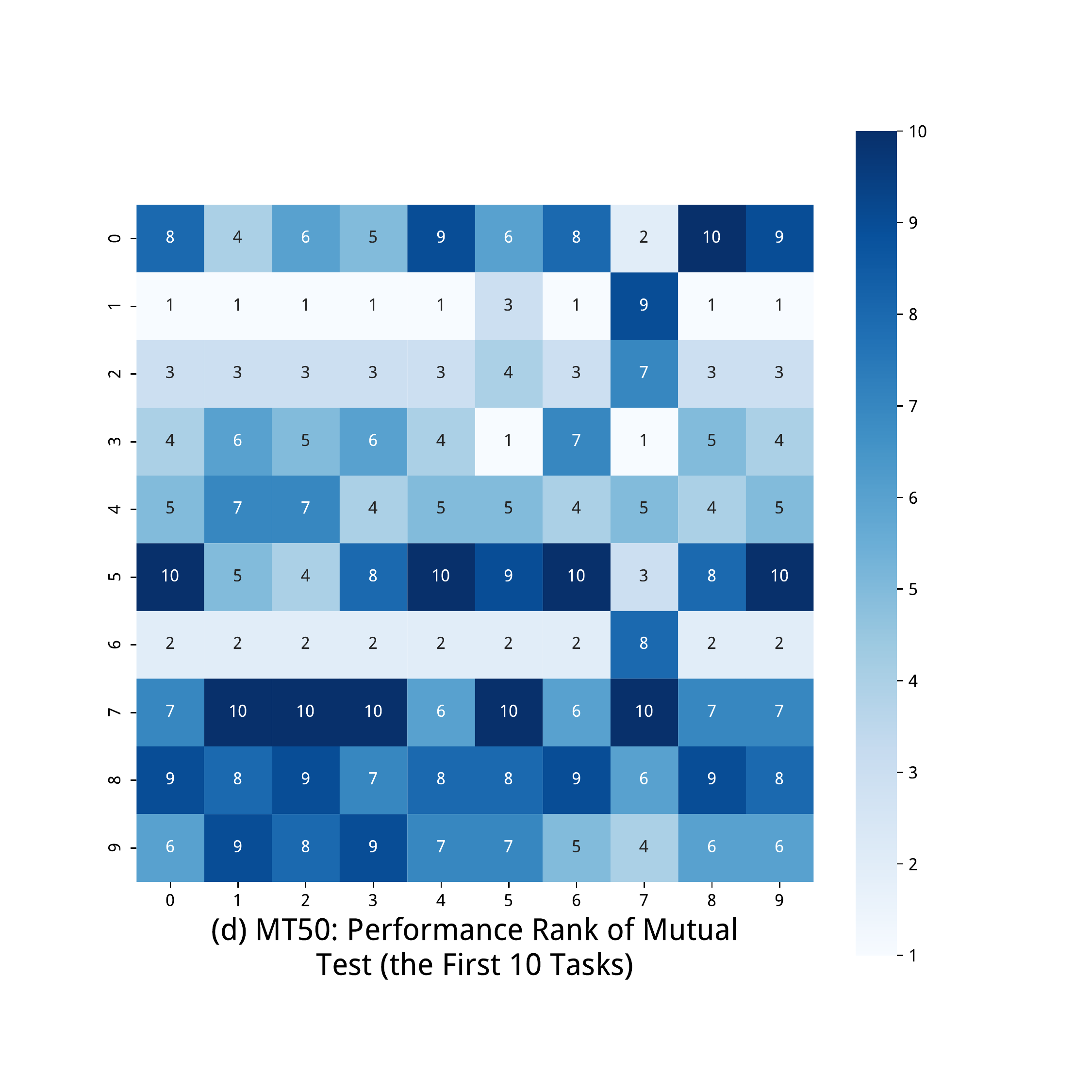} 
\caption{Comparisons of $B-I^{T\times T}$ and the performance rank matrix of mutual evaluation across tasks: (a) MT10: $B-I^{10\times 10}$; (b) MT10: $PR$ of Mutual Test; (c) MT50: $B-I^{10\times 10}$ (the First 10 Tasks); (d) MT50: $PR$ of Mutual Test (the First 10 Tasks).}
    \label{fig:matrix rank}
  \end{figure*}

We compare our CAMRL with existing state-of-the-art algorithms  on  the Gym-minigrid, Meta-world,  Atari games, Ravens, and RLBench, which are the benchmarks for multi-task RL widely used in the community \cite{chevalier2018babyai,yu2019meta,bellemare2013arcade}.

\subsection{Environments} 

 \textbf{Gym-minigrid:}\ Gym-minigrid environments \cite{chevalier2018babyai} are partially observable  grid environments consisting of tasks with increasing
complexity levels. In Gym-minigrid, the agent should find a key to open a locked door in a grid. By setting different obstacles  in the way and adding different requirements, we obtain different environments, such as DistShift, Doorkey, DynamicObstacles, etc.
Each environment can be easily tuned in terms of size/complexity, which contributes to fine-tuning the difficulty of tasks and performing curriculum learning. 
In our experiments, we first randomly select $9$ environments, each of which  owns more than $3$  tasks with different complexities. Then, we choose the task with the biggest complexity for each selected environment and form the $9$ tasks that we conduct experiments on with regard to Gym-minigrid.

\textbf{Meta-world:}\ Meta-World \cite{yu2019meta} is a  simulated benchmark for meta reinforcement learning and multi-task learning, containing $50$ tasks related to robotic manipulation. In Meta-World,  $MT1$, $MT10$, and $MT50$ are multi-task-RL  environments that have $1$, $10$, and $50$  tasks, respectively. 

The tasks for $MT10$ from Task $1$ to Task $10$ are
'Pick and place', 'Pushing', 'Reaching', 'Door opening', 'Button press', 'Peg insertion side', 'Window opening', 'Window closing', 'Drawer opening', and 'Drawer closing', respectively.

The tasks for $MT50$ from Task $1$ to Task $50$ are
'Turn on faucet', 'Sweep', 'Stack', 'Unstack', 'Turn off faucet', 'Push back', 'Pull lever', 'Turn dial', 'Push with stick', 'Get  coffee', 'Pull handle side', 'Basketball', 'Pull with stick', 'Sweep into hole', 'Disassemble nut', 'Place onto shell', 'Push mug', 'Press handle side', 'Hammer', 'Slide plate', 'Slide plate side', 'Press button wall', 'Press handle', 'Pull handle', 'Soccer', 'Retrieve plate side', 'Retrieve plate', 'Close drawer', 'Press button top', 'Reach', 'Press button top w/wall', 'Reach with wall', 'Insert peg side', 'Push', 'Push with wall', 'Pick \& place w/wall', 'Press button', 'Pick \& place', 'Pull mug', 'Unplug peg', 'Close window', 'Open window', 'Open door', 'Close door', 'Open drawer', 'Open box', 'Close box', 'Lock door', 'Unlock door', and 'Pick bin', respectively.

\textbf{Atari:}   
Atari Learning Environment (ALE) was first proposed in \cite{bellemare2013arcade}
 which includes exploration, planning, reactive play, and complex visual input \cite{espeholt2018impala}.   
 In our experiments,  we randomly select $10$ tasks with visual observations  in  ALE to perform experiments, namely, YarsRevenge, Jamesbond, FishingDerby, Venture, DoubleDunk, Kangaroo, IceHockey, ChopperCommand, Krull, and Robotank.

 \textbf{Ravens:}
 Ravens is a  vision-based robotic manipulation environment based on PyBullet. We follow \cite{zeng2020transporter} to select $10$ typical discrete-time tabletop manipulation tasks in Ravens, namely, Block-insertion, Place-red-in-green, Towers-of-hanoi, Align-box-corner, Stack-block-pyramid, Palletizing-boxes, Assembling-kits, Packing-boxes, Manipulating-rope, and Sweeping-piles. 
 Similar to \cite{zeng2020transporter}, we generate $1000$ expert demonstrations for each task by the 
 oracle provided by Ravens and use these demonstrations to perform imitation learning for each adopted baseline. After the  imitation learning phase, we train each baseline without demonstrations for another $10000$ epochs and compare the average success rate of each baseline.

\textbf{RLBench:}   RLBench is a large-scale environment designed to speed up vision-guided manipulation research. We follow \cite{liu2022auto} to test our selected baselines on $10$ typical RLBench tasks, i.e., Reach
Target,
Push
Button,
Pick And
Lift,
Pick Up
Cup,
Put Knife on
Chopping Board,
Take Money
Out Safe,
Put Money
In Safe,
Pick Up
Umbrella,
Stack
Wine, and
Slide Block
To Target.
\subsection{Baselines} 
To demonstrate the efficiency of the proposed CAMRL algorithm,  we  utilize  actor critic (AC) \cite{konda2000actor},
soft actor critic (SAC)  \cite{HaarnojaZAL18}, mastering rate based online curriculum learning (MRCL) \cite{willems2020mastering}, YOLOR \cite{wang2021you}, distral  \cite{teh2017distral}, gradient surgery \cite{yu2020gradient}, and soft module \cite{yang2020multi}  as our baselines for comparison. The criteria of choosing baselines depend on whether they are commonly-used or state-of-the-art methods. 

Among the above algorithms, regarding the mastering rate based online curriculum learning (MRCL),
\cite{willems2020mastering}
made the assumption that the good next tasks are those that are learnable but not learned thoroughly yet,
and  thus introduced a new algorithm based on the notion of mastering rate so as to  choose which task to train next in an online manner. Due to the fact that the rewards of tasks in gym-minigrid are pretty sparse, we customize a curriculum by the size of grid for each task,  apply MRCL to switch the sub-tasks in each curriculum from the micro perspective, and utilize other baselines to perform multi-task or single-task training for all the tasks in  the macro aspect.

\subsection{Evaluation Metric}
We adopt the reward metric in the Gym-minigrid, meta-world, and ALE environments and the success rate metric in the Ravens and RLBench environments.
Although with regard to the meta-world environments, the success rate has been utilized in \cite{yu2020gradient,yang2020multi}, this evaluation metric can be zero very often, which leads to sparse feedback and is bad for comparing the difficulties of training different tasks. Besides,  \cite{yu2020gradient,yang2020multi} displayed performance of their  algorithms with the averaged  success rate of all tasks instead of  individual  success rates  for each task, which makes the  success rate less informative than the reward metric for each task that we adopt in this paper. Last but not the least, the  simulation steps in \cite{yu2020gradient}, the number of samples in \cite{yang2020multi}, and the number of epochs in this paper are not in the same magnitude, which makes our reference for the success rate in \cite{yu2020gradient,yang2020multi} less meaningful.
 As a result, in the meta-world environment, we   use  reward  instead of the averaged success rate to formulate the evaluation metric, the policy loss in  $I_{mul}$, and the composite loss terms.



\subsection{Implementation Details}\ 
We adopt the same network structure for each single actor network and single critic network in each baseline. For soft module and gradient surgery, we follow the same parameters as in their public codes. For distral, we follow \cite{teh2017distral} to set $\alpha=0.5$ and  use a set of $9$ hyper-parameters $(1 / \beta, \epsilon) \in\left\{3 \cdot 10^{-4}, 10^{-3}, 3 \cdot 10^{-3}\right\} \times\left\{2 \cdot 10^{-4}, 4 \cdot 10^{-4}, 8 \cdot 10^{-4}\right\}$  for the
entropy costs and the initial learning rate. Hyper-parameters with the optimal result are utilized with regard to the distral algorithm.

Concerning our CAMRL, we set $a=1000,b=\frac{1}{40},c=2,d=200,\mu_1=\mu_2=0.01$, and $radius=0.05$ in all tasks.
At the end of every epoch, instead of performing mutual test for each pair of tasks which is computationally-inefficient, we  1) randomly select two tasks $i$ and $j$; 2) test the performance of $SAC_i$ on task $j$ and the performance of $SAC_j$ on task $i$; 3) update $p_{i,j}$ and $p_{j,i}$; and 4) repeat the above mutual test procedure for $3$ times.
Besides, to avoid  the abrupt increase in  loss  when switching to the curriculum-based AMTL mode, the actual loss  for task $t$ is set to be $\mathcal{L}(w_t)$ plus 0.01 times the objective  in Eq. (\ref{eq:7}). In addition, the proposed algorithms are run over $5$ seeds  to report the averaged results. 

\subsection{Results}
\textbf{Results on gym-minigrid.}
As shown in Table \ref{tab:minigrid}, among the cooperation between mastering rate based online  curriculum learning (MRCL) \cite{willems2020mastering}  and the single-task actor critic/distral/gradient surgery/CAMRL algorithm,  CAMRL achieves the best overall performance over the $9$ minigrid tasks, without showing obvious signs of negative transfer.

\textbf{Results on MT10 and MT50.}
As can be seen in Table \ref{tab:mt10}, 
  CAMRL   significantly beats all baselines over $6$ tasks out of the $10$ tasks in MT10. The second-best algorithm overall is   single SAC  and is obviously worse than CAMRL in $7$ tasks out of all  $10$ tasks.
In Table  \ref{tab:mt50_whole2}, the best and second-best algorithms overall are  CAMRL and soft module, respectively. The latter is at least 20\% worse than  CAMRL in $32$ tasks out of the total $50$ tasks and at least 20\% better than CAMRL in $8$ tasks out of all   tasks. To sum up, in existing experiments, our CAMRL works well  whether the task scale is large or small. Moreover, it seldom exposes serious negative transfer, and can sometimes contribute greatly to the tasks that are hard to train for other baselines.

\textbf{Results on Atari, Ravens, and RLBench.}\ 
See the performance of all algorithms on the selected Atari, Ravens, and RLBench tasks  in Tables \ref{tab:atari}, \ref{tab:ravens}, and \ref{tab:rlbench}.
From Table \ref{tab:atari}, \ref{tab:ravens}, and \ref{tab:rlbench}, we can see that our CAMRL algorithm ranks first in $6$ tasks out of the total selected $10$ Atari  tasks, $8$ tasks out of the total selected $10$ Ravens  tasks, and $9$ tasks out of the total selected $10$ RLBench  tasks.
No significant negative transfer is found when performing CAMRL in all these tasks.

\textbf{Analysis on the transfer matrix $B$.}
As stated in Section \ref{algorithm}, $B$ is   a $T \times T$ asymmetric matrix representing
the number of  transfers between tasks.  
Figure \ref{fig:matrix rank}(a) visualizes the $B-I^{10\times 10}$ matrix where $I_{10\times 10}$ is the identity matrix in $\mathbb{R}^{10\times 10}$, and Figure \ref{fig:matrix rank}(c)   visualizes the $B-I^{10\times 10}$ matrix of the first $10$ tasks in MT50.

To discover some in-depth rules related to $B$, we calculate $PM\in \mathbb{R}^{T\times T}$,   the  performance matrix of mutual evaluation across tasks, and compare its variant $PR$, a performance rank matrix, with $B-I^{10\times 10}$:  First, test the average performance of the single SAC net trained for task $s$ over $10k$ episodes on task $t$, and regard it as $PM_{s,t}$ ($s,t\in [T]$) (the evaluation is run for $20$ episodes); next, for each task $t\in [T]$, rank  $PM_{s,t} (s\in [T])$ by the order from minimum to maximum and regard the rank as the $t$-th column of the performance rank matrix $PR\in \mathbb{R}^{T\times T}$.
The visualization of $PR$ for MT10 is shown in Figure \ref{fig:matrix rank}(b), and Figure \ref{fig:matrix rank}(d) displays the $PR$ for the first $10$ tasks out of the $50$ tasks in MT50. The bigger the number in $PR_{s,t}$, the better evaluation performance that  the SAC network originally trained for task $s$ has on task $t$.
As   shown in Figure \ref{fig:matrix rank}, when  $PR_{s,t}$ is small enough, say no larger than $2$ (see Figure \ref{fig:matrix rank} (b)(d)),   $B_{s,t}-I_{s,t}^{T\times T}$ is also obviously smaller than other elements  (see Figure \ref{fig:matrix rank} (a)(c)). This, to some extent  from the opposite side, confirms our previous demand that the better evaluation performance  on other tasks, the larger number of transfers on those tasks.

\textbf{Incorporation of prior knowledge.} To show the capability  of incorporating  prior knowledge for CAMRL, we perform extra experiments that formulate a new differentiable ranking loss for tasks where the relative magnitudes of the task difficulty are readily apparent in part. Specifically, we follow \cite{metaworld}, \cite{metaworld}, \cite{c.elmohamed},  \cite{zeng2020transporter}, and \cite{liu2022auto} to obtain the public performance of existing state-of-the-arts methods for MT10, MT50, Atari,  Ravens, and RLBench, respectively. Then we use the relative ranking of the public performance to formulate a new $tanh$-based differentiable ranking loss and incorporate it with Eq. (\ref{eq:7}), in the hope that for task $t\in [T]$, if task $i$ is easier to train, that is, with  better public performance, then task $t$ tends to perform   a smaller number of transfers  to task $i$. The results of adding this new differentiable ranking loss term are shown in Tables \ref{tab:mt10_prior}, \ref{tab:mt50_prior}, \ref{tab:atari_prior},  \ref{tab:ravens_prior}, and \ref{tab:rlbench_prior}. 
We can see that after adding the new loss term, the performance of CAMRL beats
its original version over $5$ tasks out of the $10$ tasks in MT10, $38$ tasks out of the $50$ tasks in MT50, $8$ tasks out of the $10$ tasks in Atari,  $7$ tasks out of the $10$ tasks in Ravens, and $9$ tasks out of the $10$ tasks in RLBench,  which demonstrates CAMRL's capability of incorporating  prior knowledge   as well as the great performance of the differentiable ranking loss in CAMRL.

\begin{figure}
    \centering
    \includegraphics[width=0.7\linewidth]{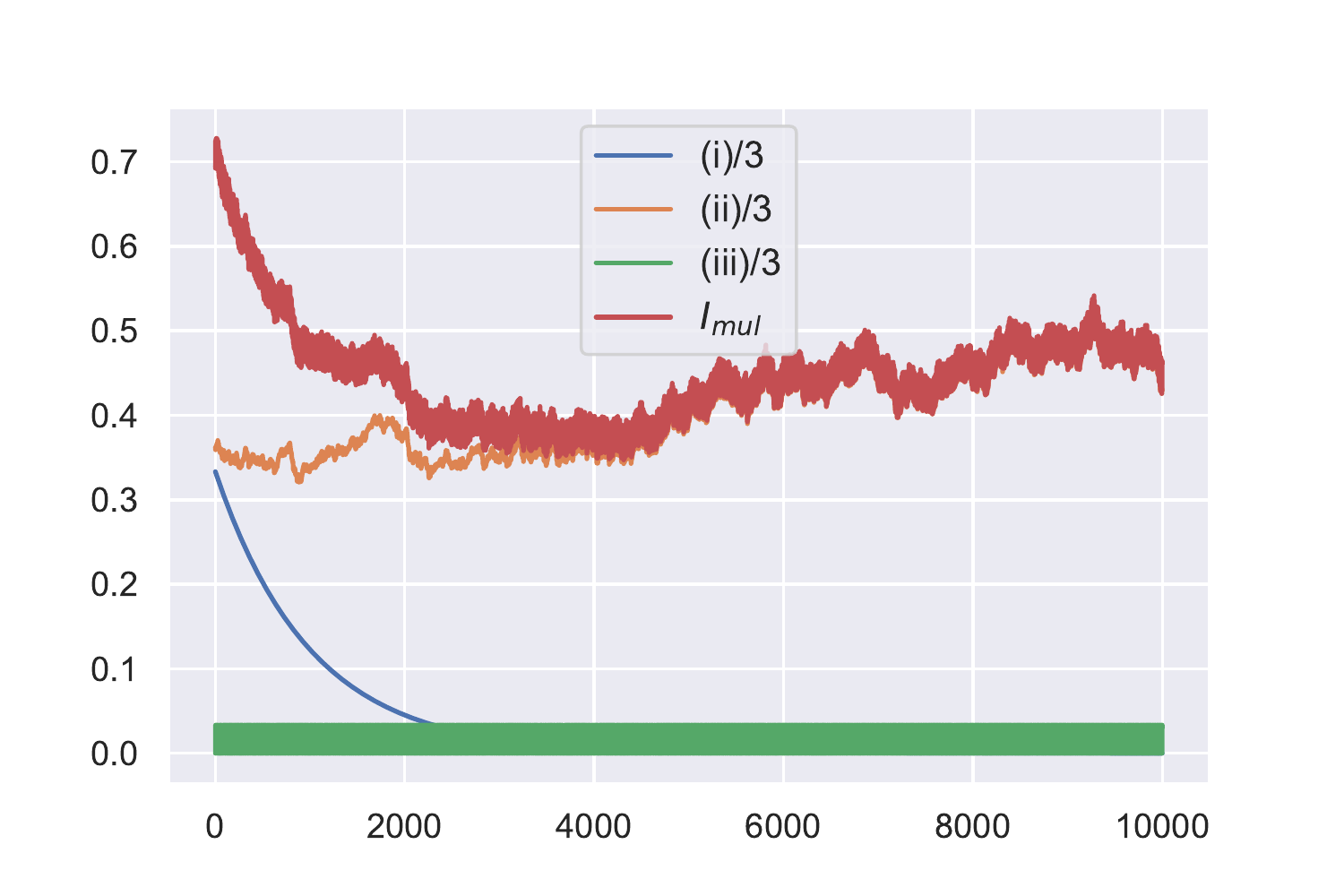}
    \caption{Changes of terms in $I_{mul}$ for MT10.}
    \label{fig:imul}
\end{figure}

\textbf{Changes of each term in $I_{mul}$.} 
To demonstrate the switching process of CAMRL's training mode and visualize the magnitude change of each term in $I_{mul}$, we plot the changes of  $I_{mul}$ and its three terms for CAMRL in the MT10 environment. From Figure \ref{fig:imul} we can see that
the magnitude of each term in $I_{mul}$ does not differ  much and $I_{mul}$ keeps ranging from $0.2$ to $0.8$, which does not show  abnormal fluctuation and  magnitude much.

\textbf{Hyper-parameter analysis.}
The results of the hyper-parameter analysis for our CAMRL on MT10 are shown in Figure \ref{fig:para}.
We first set $ \lambda_0=1,  \mu_1=\mu_2=\lambda_1=\lambda_2=\lambda_3=\lambda_4=0.01, a=1000, b=\frac{1}{40}, c=2, d=200$, and $radius=0.05$. And the evaluation metric is defined as the average reward of all $10$ tasks over the first $10k$ episodes by the CAMRL algorithm. Then, we

{\bf (i)} fix $\mu_2$, $\lambda_1$, $\lambda_2$, $\lambda_3$, $\lambda_4$, $a$, $b$, $c$, $d$, and $radius$.  Set parameter $\mu_1=0, 0.001, 0.002, 0.005, 0.01, 0.02, 0.05, 0.1, 0.2, 0.5, 1$, respectively. The results under different values of $\mu_1$ are shown in Figure \ref{fig:para} (line $\mu_1$);
    
    {\bf (ii)} repeat the same procedure for $\lambda_i (i\in [4])$, $a$, $\frac{1}{b}$, $c$, $d$, and $\mu_2$ as that for $\mu_1$.

From Figure \ref{fig:para}(a) we can see that, as for $\mu_i (i\in [2])$ and $\lambda_i (i\in [4])$,  when all of these hyper-parameters are around $0.01$, decreasing one parameter among them to $0$ may hurt the performance. In the meantime, if one of the parameters is  increased too much, the great performance of CAMRL may not be guaranteed, which may result from the abrupt change of loss when switching the training mode and from the exceeded negative transfer.
In summary, according to the above experimental evaluation, as for the time-invariant design of $\lambda_i (i\in [4])$,
a reasonable setting, e.g., $\lambda_i\approx 0.01,\mu_j\approx 0.01 \text{ for } i\in [4],j\in [2] $, is indeed helpful in improving SAC's performance, compared with the scenarios that some hyper-parameters are set to zeros. Furthermore, as shown in Figure \ref{fig:para},  when we automatically adjust $\lambda_i (i\in [4])$ based on each term's uncertainty, the performance of CAMRL outweighs most configurations of $\lambda_i (i\in [4])$.

With regard to hyper-parameters 
 $a$, $\frac{1}{b}$, $c$, and $d$, as can be seen in Figure \ref{fig:para}(b), our CAMRL is
 the most sensitive to $d$, the co-efficient of the inner term in the $\tanh$ ranking function, which strongly demonstrates the effectiveness of our customized differentiable   
ranking mechanism. Moreover,
increasing 
$a$, $\frac{1}{b}$, and $c$ to an appropriate value is also beneficial to CAMRL's performance and does not show a significant sign of performance deterioration due to the magnitude problem of  terms in the index $I_{mul}$.

\begin{remark}
In fact, due to our elaborate design, each term in $I_{mul}$ normally will not exceed $\frac{1}{3}$. And if one of the terms becomes very small and fails to work out in the index, the other terms will still function so that CAMRL's performance will not deteriorate too quickly when performing a hyper-parameter analysis on $a$, $b$, and $c$.
\end{remark}

\textbf{Ablation study.}
For the ablation study, on tasks MT10, we test our method   in the lack of the mode switching component (i.e., entirely performing curriculum learning), and when setting $\frac{1}{a}/b/c/d/\lambda_i (i=2,3,4)$ to be $0$, respectively. Also, we test the performance of AMRL which lacks the mode switching mechanism and multiple ranking functions compared with our CAMRL.  As shown in Table \ref{tab:ablation study},  the lack of any component of our CAMRL and  setting $\frac{1}{a}/b/c/d/\lambda_i (i=2,3,4)$ to be $0$ or positive infinity (so that the corresponding term in $I_{mul}$ is $0$)  bring varying degrees of reduction in algorithm performance, which consistently demonstrates the precision of algorithm design and the importance of tacit cooperation 
among different components. The obvious performance deterioration  when only performing AMRL and when lacking the mode switching component specifically shows the effectiveness and efficiency of our  design of the index $I_{mul}$ and the customized differentiable ranking functions.

\section{Discussion}

In this section, we discuss the advantages of  CAMRL over existing MTRL algorithms, some clarifications of CAMRL, and direct ways to improve CAMRL.

\subsection{Advantages over Existing MTRL Algorithms}
When there exist many tasks, it is difficult to find common traits  owned by  all tasks, and thus, the distillation-like algorithms may perform poorly.
On the contrary, except for distilling common traits of all tasks by a large network,
our $B$ matrix  probes the transfer relationship between every two tasks and enables us to perform asymmetric transfer  between tasks  so that CAMRL will not be troubled by this issue.

For  multi-task learning with a  modular paradigm,   negative transfer can be  serious since the task relationship
in the early period is  incorrect and the bad influence may accumulate. 
Moreover, it may take a long time to approximately capture the correct task relationship and by the time the relationship is well learned, while the effect caused by negative transfer might have already been irreversible. On the contrary, by initializing   $B$   as a unit matrix (no transfer at the beginning), adding constraints to  $B$, and switching between various training modes, CAMRL can 
not only mitigate negative 
transfer, but also  learn a good $B$ at a relatively early stage, allowing the training on difficult tasks to be positively affected by  other tasks.

 
We summarize the advantages of CAMRL as five-fold. 
First, CAMRL is flexible: 
\textbf{(i)} CAMRL can freely switch between parallel single-task training  and curriculum-based  AMTL  modes according to the indicator related to the learning progress.
\textbf{(ii)} The sub-components of CAMRL have variants
and can  be paired with different RL baseline algorithms and training modes. CAMRL does not impose restrictions on the network architecture and learning mode.
For example, $W$ could be parameters of the whole network. When there is a large number of tasks, we could share the principle part of the whole network and let $W$ be only a small subset of the network in order to save  RAM. 
Second, CAMRL is promising in mitigating negative transfer. For one thing, by considering the $1$-norm constraint on   $B$ and applying   Frank-Wolfe   to meet the constraint, CAMRL can avoid excessive transfer to some extent. For another,
by customizing and utilizing indicators about the learning progress and the performance  of mutual tests across tasks, CAMRL can alleviate negative transfer. 
Third, the loss function  of CAMRL can take information from multiple aspects to improve the efficiency of RL tasks' training. Specifically,
the loss function in Eq. \eqref{eq:5} considers the positivity and sparsity of the transfer matrix, the training difficulty
for different tasks,   the  performance   of mutual tests, and the similarity between every two tasks. Thanks to our customized ranking loss which enables to absorb partial ranking information, CAMRL could take full advantage of various  existing prior knowledge and training factors, regardless of their amount.
Fourth, CAMRL   eliminates the need for laborious hyper-parameter analysis.  Specifically, CAMRL allows the hyper-parameters in the composite loss to dynamically auto-adjust and regards the auto-adjusting process as a multi-task learning problem. An uncertainty-based multi-task learning method is utilized to update the hyper-parameters automatically.
Fifth, CAMRL can be auto-adapted to a new task 
without affecting original tasks by changing $B$ to $B^{\prime}= [B,\ (0,\dots,0)^{\top}; (0,\dots,0),\  1 ]$.
 
\subsection{Some Clarifications}\label{sec:diss}

\textbf{Computation amount of mutual evaluation.}  If the mutual evaluation between tasks requires a big amount of computation, we could reduce the frequency of mutual evaluation between tasks and reuse the evaluation results between each evaluation. Besides, in each epoch, we could just randomly select a few tasks for mutual evaluation and only conduct partial ranking with regard to the ranking loss.
    
\textbf{Complexity of the loss term.} Although our customized loss contains multiple terms which might lead to strenuous hyper-parameter tuning, we adapt the automatic hyper-parameter adjustment technology in \cite{kendall2018multi} to handle the above issue. Moreover, in the future, it might be valuable to discover other factors that make a more significant difference to CAMRL and simplify the loss term by replacing the existing factors in the loss with the most 
   remarkable factors.
   
    
\textbf{Performances of the trade-off mechanism among  differentiable loss functions.}
    Although we adopt differentiable loss functions in order to keep the ranking alignment between $B_{t,i} (i\in [T])$ and $p_{t,i}$(\text{or }$\mathcal{L}(w_i))$, there are multiple terms in the loss and thus we have to make a compromise between different terms. As a result, it is difficult to achieve perfect alignment concerning each ranking loss, and to mitigate negative transfer we mainly hope that $B_{t,i}$ can be small when $p_{t,i}$ and $\mathcal{L}(w_i)$ are small enough. Actually, as shown in Figure \ref{fig:matrix rank}, when  $PR_{s,t}$ is small enough, say no larger than $2$,   $B_{s,t}-I_{s,t}^{T\times T}$ is also obviously smaller than other elements, which meets our expectation for relieving serious negative transfer.

\textbf{Details of integrating MRCL with the baselines.}
    Since the tasks we select from Gym-minigrid are of the biggest size or complexity,  directly overcoming these tasks  would be very difficult. Therefore, to lower the barrier of tasks' learning,  we follow the mastering rate based online curriculum learning (MRCL) algorithm \cite{willems2020mastering} to perform curriculum learning for each selected task.
    Specifically, 
    for each selected task $i$, we take  task $i$ and
    other easier tasks in Gym-minigrid with the same property as task $i$ as the curriculum of task $i$. During the training phase of each task $i$, 
    we adopt MRCL  to learn the curriculum of task $i$ as quickly as possible, with the learning algorithm set to be our selected baseline. At each training step, MRCL  adaptively adjusts the next task in the  curriculum of task $i$ for training according to the dynamic attention for each task,
    under the assumption that the good next tasks are the ones which are learnable but not learned yet. Thanks to the delicate attention mechanism of MRCL, the learning algorithm could overcome the selected tasks in Gym-minigrid little by little instead of being stuck somewhere for a long time.
    
 \textbf{Configuration method of hyper-parameters.} (i)
    For hyper-parameters $a, b$, and $c$, we adjust them according to the magnitude of the variables in each item of the index $I_{mul}$, with the hope that the composite magnitude of each item in $I_{mul}$ does not differ too far. (ii)
     For the hyper-parameter $d$, we plot various ranking functions with different $d$ and select $d$ with a relatively smooth ranking function as well as  being as small as possible, since a small $d$ could contribute to the convergence speed of vanilla Frank-Wolfe during the optimization of $b_t^o$. (iii)
     For hyper-parameters $\lambda_i (i\in \{0,1,2,3,4\})$, we automatically adjust them according to an uncertainty-based multi-task learning method in \cite{kendall2018multi}.

\subsection{Potential Improvements}
Below we list several directions for CAMRL to be improved, so as to illuminate further incorporation of curriculum-based AMTL into RL.

\begin{itemize}
    \item Consider more auto-tuning approaches of hyper-parameters $\mu_j (j=1,2)$ and  $\lambda_j (j=1,2,3,4)$, as well as  allowing $\mu_j$ to  be the time-varying function of the  learning progress and other factors.
    
    \item Enrich the loss by integrating the intermediate difficulty, diversity, surprise and energy  \cite{portelas2020automatic}, etc. 
    
    \item Design techniques to select the optimal subset of all the  networks' parameters as $W$. 
    \item Transform $B$ into a non-linear function and update it with theoretical guidance.
    \item Apply the sliding window, the exponential discounted factor, the appropriate entropy, or other tricks to 
    mitigate the non-stationarity during the AMTL training.
\end{itemize}
 
\section{Conclusions and Future Work}

In this paper, we proposed a novel multi-task reinforcement learning algorithm, called CAMRL (curriculum-based asymmetric multi-task reinforcement learning). 
CAMRL switches the training mode between parallel single-task and curriculum-based AMTL. 
CAMRL employs a composite loss function to reduce negative transfer in multi-task learning. 
Apart from regularizing tasks' outgoing transfer and network weights' similarity, we introduced three differentiable ranking functions into the loss to incorporate various prior knowledge flexibly. An alternating optimization with   Frank-Wolfe   has also been utilized to optimize the loss,
and an uncertainty-based automatic adjustment mechanism of hyper-parameters has been adopted to eliminate laborious hyper-parameter analysis.
We have conducted extensive experiments to confirm the effectiveness of  CAMRL  and have analyzed its flexibility  from various perspectives. 

In the future, we plan to   study CAMRL with more theoretical insights and consider its  potential improvements  listed in  the Discussion Section, such as  incorporating more prior knowledge into the loss, designing the non-linear version of the transfer matrix, and overcoming the non-stationarity during the AMTL training. 


\ifCLASSOPTIONcaptionsoff
  \newpage
\fi
 
\bibliography{main}
\bibliographystyle{abbrv}

\begin{IEEEbiography}
[{\includegraphics[height=1.16in,clip,keepaspectratio]{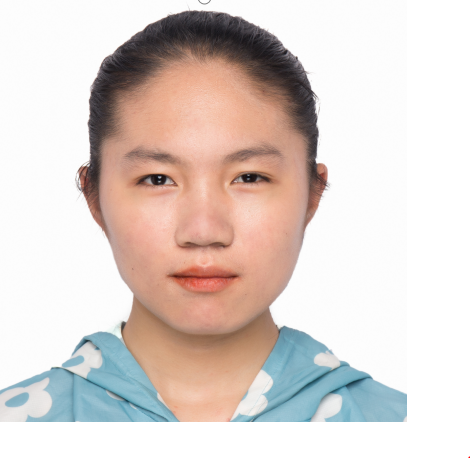}}]
{Hanchi Huang} received her bachelor's degree from Shanghai Jiao Tong University in 2021. She is now a graduate student in the School of Computer Science and Engineering, Nanyang Technological University, Singapore.
Her research interests include theory and algorithms for reinforcement learning, combinatorial optimization, and recommendation system.
\end{IEEEbiography}

\begin{IEEEbiography}
[{\includegraphics[height=1.16in,clip,keepaspectratio]{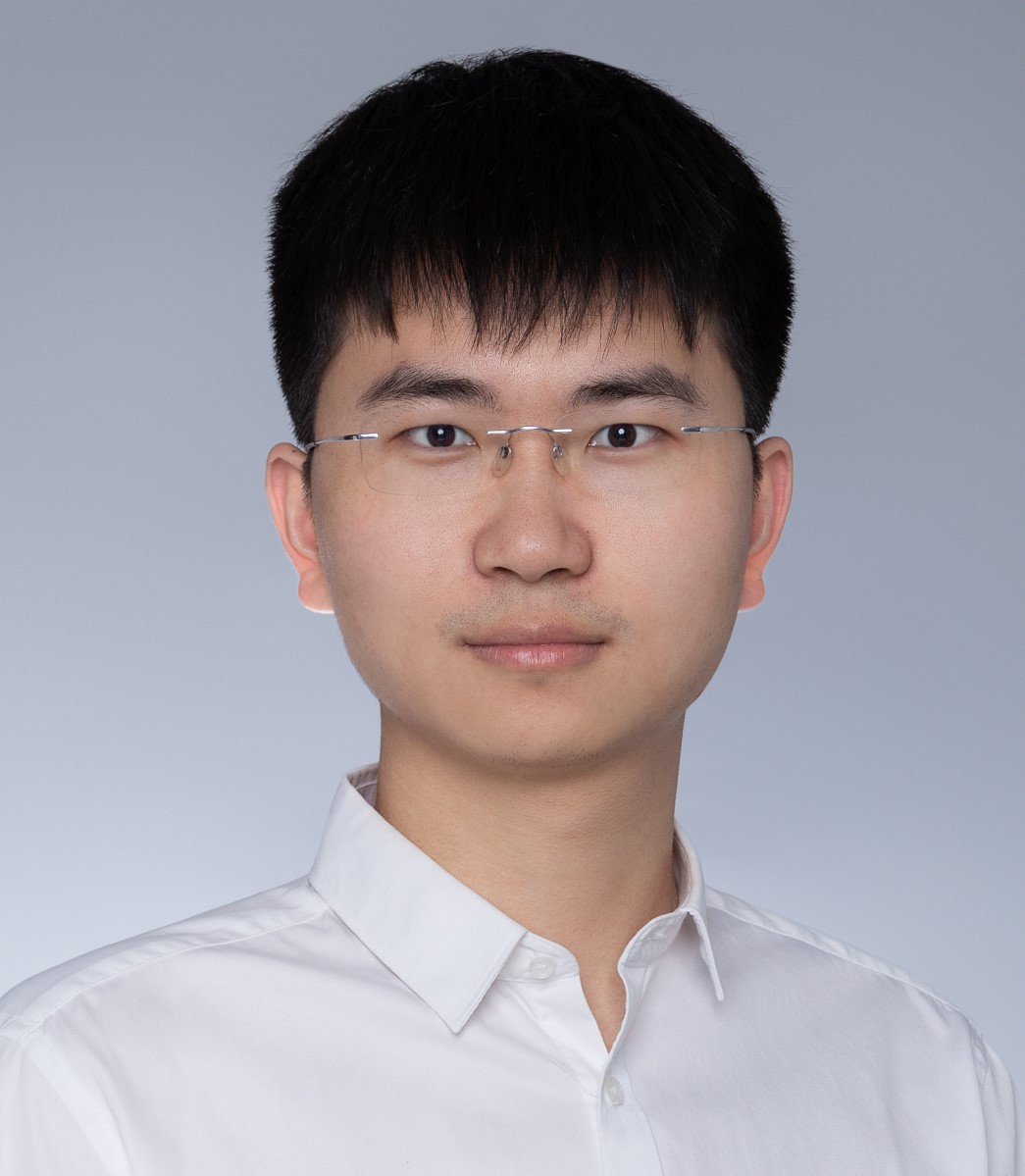}}]
{Deheng Ye} finished his Ph.D. from the School of Computer Science and Engineering, Nanyang Technological University, Singapore, in 2016. He is now a Principal Researcher and Team Manager with Tencent, Shenzhen, China, where he leads a group of engineers and researchers developing large-scale learning platforms and intelligent AI agents. He is interested in applied machine learning, reinforcement learning, and software engineering. He has been serving as a PC/SPC for NeurIPS, ICML, ICLR, AAAI, and IJCAI.
\end{IEEEbiography}

\begin{IEEEbiography}
[{\includegraphics[height=1.16in,clip,keepaspectratio]{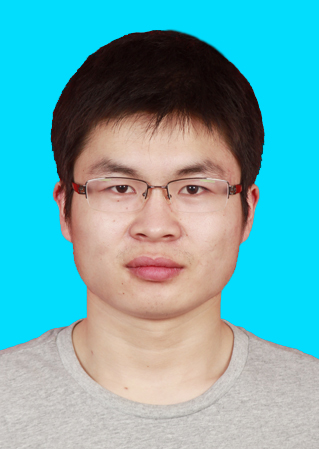}}]
{Li Shen} received his Ph.D. from the School of Mathematics, South China University of Technology, in 2017. He is now a Research Scientist at JD Explore Academy, China. Previously, he was a Research Scientist at Tencent, China. His research interests include theory and algorithms for large scale convex/nonconvex/minimax optimization problems, and their applications in statistical machine learning, deep learning, reinforcement learning, and game theory. 
\end{IEEEbiography}

\begin{IEEEbiography}
[{\includegraphics[height=1.16in,clip,keepaspectratio]{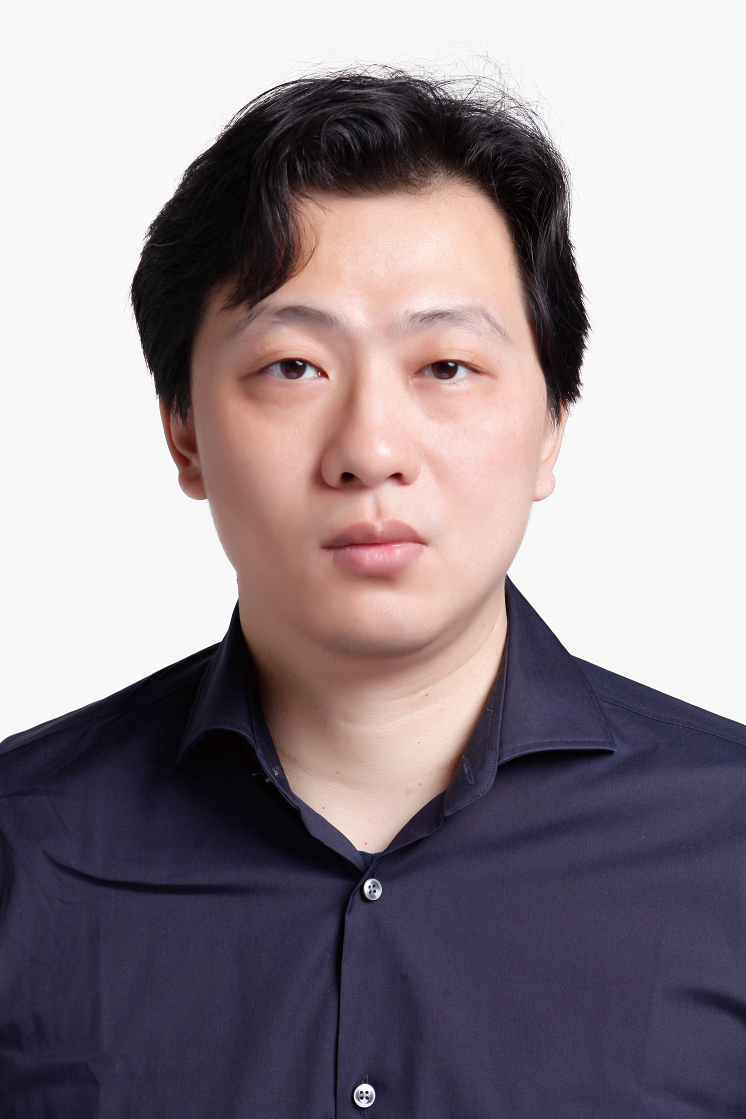}}]{Wei Liu} (M'14-SM'19) is currently a Distinguished Scientist of Tencent and the Director of Ads Multimedia AI at Tencent Data Platform. Prior to that, he has been a research staff member of IBM T. J. Watson Research Center, USA. Dr. Liu has long been devoted to fundamental research and technology development in core fields of AI, including deep learning, machine learning, computer vision, pattern recognition, information retrieval, big data, etc. To date, he has published extensively in these fields with more than 270 peer-reviewed technical papers, and also issued 23 US patents. He currently serves on the editorial boards of IEEE TPAMI, TNNLS, IEEE Intelligent Systems, and Transactions on Machine Learning Research. He is an Area Chair of top-tier computer science and AI conferences, e.g., NeurIPS, ICML, IEEE CVPR, IEEE ICCV, IJCAI, and AAAI. Dr. Liu is a Fellow of the IAPR, AAIA, IMA, RSA, and BCS, and an Elected Member of the ISI.    
\end{IEEEbiography}

\newpage

\balance
\section{Appendix}

\begin{table*}
\centering
\scriptsize
\caption{Reward before and after the incorporation of prior knowledge for  MT10.} \label{tab:mt10_prior}
\begin{tabular}{lccc}
\toprule
               & Ranking of  performance by soft-module \cite{yang2020multi}  & CAMRL w/o  prior success rate & CAMRL w/ prior  success rate \\\midrule
Pick and place     & 9                                         & \textbf{47235.31}                                & 42798.2                         \\
Pushing            & 8                                & 213.27                  

& \textbf{227.94 }                         \\
Reaching           & 2                                         & \textbf{-50.36  }                                & -51.28                          \\
Door opening       & 6                                         & 143.74                                  & \textbf{147.64 }                         \\
Button press       & 4                                         & 1395.77                                 & \textbf{1449.87}                         \\
Peg insertion side & 10                                        & \textbf{52.91 }                                  & 50.03                           \\
Window opening     & 3                                         & \textbf{15683.29 }                               & 14948.5                         \\
Window closing     & 7                                         & 23.64                                   & \textbf{26.97}                           \\
Drawer opening     & 5                                         & 1329.39                                 & \textbf{1403.84}                         \\
Drawer closing     & 1                                         & \textbf{622.74}                                  & 618.76                         \\\bottomrule
\end{tabular}

\end{table*}

\begin{table*}
\centering
\scriptsize
\caption{Reward before and after the incorporation of prior knowledge for  MT50.} \label{tab:mt50_prior}
\begin{tabular}{lccc}
\toprule
               & Ranking of  performance by MT-SAC of meta-world benchmark \cite{metaworld}  & CAMRL w/o  prior success rate & CAMRL w/ prior  success rate \\\midrule
Turn on faucet          & 1                                         & 7644.89                                 & \textbf{7928.39 }                        \\
Sweep                   & 37                                        & -10.31                                  & \textbf{-6.38}                           \\
Stack                   & NA (not found)                                      & \textbf{-42.32}                                  & -43.21                          \\
Unstack                 & NA (not found)                                      & -31.26                                  & \textbf{-30.48}                          \\
Turn off faucet         & 1                                         & 1253.73                                 & \textbf{1308.41}                         \\
Push back               & 34                                        & -27.74                                  & \textbf{-26.37}                          \\
Pull lever              & 36                                        & -31.84                                  & \textbf{-30.37}                          \\
Turn dial               & 1                                         & \textbf{-27.69}                                  & -28.36                          \\
Push with stick         & 37                                        & 1376.83                                 & \textbf{1427.37}                         \\
Get coffee              & 31                                        & -52.93                                  & \textbf{-49.31 }                         \\
Pull handle side        & 1                                         & \textbf{-48.93}                                  & -49.76                          \\
Basketball              & 37                                        & 16394.92                                & \textbf{17389.21}                        \\
Pull with stick         & 37                                        & -28.92                                  & \textbf{-26.43}                          \\
Sweep into hole         & NA (not found)                                      & 16.74                                   & \textbf{17.38}                           \\
Disassemble nut         & 37                                        & \textbf{1873.93}                                 & 1793.35                         \\
Place onto shell        & NA (not found)                                      & 274.01                                  & \textbf{304.74}                          \\
Push mug                & NA (not found)                                      & 638.91                                  & \textbf{652.48}                          \\
Press handle side       & 1                                         & 24.18                                   & \textbf{25.31}                           \\
Hammer                  & 26                                        & -60.93                                  & \textbf{-58.42}                          \\
Slide plate             & 1                                         & 425.64                                  & \textbf{445.31}                          \\
Slide plate side        & 24                                        & \textbf{-23.54}                                  & -24.01                          \\
Press button wall       & 1                                         & -27.73                                  & \textbf{-26.46}                          \\
Press handle            & 1                                         & -49.18                                  & \textbf{-47.54}                          \\
Pull handle             & 1                                         & 46.84                                   & \textbf{47.14 }                          \\
Soccer                  & 31                                        & \textbf{485.03 }                                 & 472.77                          \\
Retrieve plate side     & 1                                         & 127.94                                  & \textbf{130.58}                          \\
Retrieve plate          & 1                                         & \textbf{-42.65 }                                 & -44.51                          \\
Close drawer            & 1                                         & 428.93                                  & \textbf{ 441.93 }                         \\
Press button top        & 1                                         & -29.84                                  & \textbf{-24.18 }                         \\
Reach                   & 1                                         & -23.47                                  & \textbf{-20.45 }                         \\
Press button top w/wall & 1                                         & -41.84                                  & \textbf{-38.53 }                         \\
Reach with wall         & 1                                         & 897.42                                  &\textbf{ 910.58 }                         \\
Insert peg side         & 25                                        & \textbf{12.93  }                                 & 10.56                           \\
Push                    & 30                                        & \textbf{573.94}                                  & 570.53                          \\
Push with wall          & 35                                        & 23.81                                   & \textbf{ 30.51  }                         \\
Pick \& place w/wall    & 37                                        & -33.27                                  & \textbf{-30.56 }                         \\
Press button            & 1                                         & 847.93                                  & \textbf{856.25 }                         \\
Pick \& place           & 37                                        & -23.15                                  & \textbf{-21.16}                          \\
Pull mug                & NA (not found)                                      & -3.03                                   & \textbf{-2.74    }                       \\
Unplug peg              & 29                                        & -52.94                                  & \textbf{-48.61}                          \\
Close window            & 1                                         & -21.19                                  & \textbf{-20.68 }                         \\
Open window             & 22                                        & \textbf{-27.36 }                                 & -29.63                          \\
Open door               & 22                                        & -62.27                                  & \textbf{-58.64 }                         \\
Close door              & 1                                         & \textbf{-23.01 }                                 & -25.96                          \\
Open drawer             & 27                                        & -60.03                                  & \textbf{-56.61}                          \\
Open box                & 33                                        & -27.09                                  & \textbf{-24.45 }                         \\
Close box               & 28                                        & -47.15                                  & \textbf{-44.58 }                         \\
Lock door               & 1                                         & -45.39                                  & \textbf{-40.99}                          \\
Unlock door             & 1                                         & \textbf{-33.06 }                                 & -36.81                          \\
Pick bin                & 37                                        & -42.94                                  & \textbf{-40.51  }                           \\\bottomrule
\end{tabular}

\end{table*}
\begin{table*}
\centering
\scriptsize
\caption{Reward before and after the incorporation of prior knowledge for Atari.}\label{tab:atari_prior}


\begin{tabular}{lcccc}
\toprule
          & Ranking of  performance in  the Atari Learderboard & CAMRL w/o prior reward & CAMRL w/ prior reward \\\midrule
YarsRevenge    & {\color[HTML]{24292F} 1}          & 1637.94                          & \textbf{1684.8}                   \\
Jamesbond      & NA (not found in the Atari leaderboard \cite{c.elmohamed})                          & \textbf{8.23 }                            & 7.96                     \\
FishingDerby   & {\color[HTML]{24292F} 7}          & -11.41                           & \textbf{-7.43  }                  \\
Venture        & {\color[HTML]{24292F} 5}          & 0.22                             & \textbf{ 0.24}                     \\
DoubleDunk     & NA (not found)     & -0.89                            & \textbf{-0.75}                    \\
Kangaroo       & {\color[HTML]{24292F} 3}          & 5.74                             &\textbf{ 5.94 }                    \\
IceHockey      & {\color[HTML]{24292F} 8}          & \textbf{-0.49 }                           & -0.52                    \\
ChopperCommand & {\color[HTML]{24292F} 2}          & 351.68                           & \textbf{374.95  }                 \\
Krull          & {\color[HTML]{24292F} 4}          & \textbf{9.75 }                            & 9.49                     \\
Robotank       & {\color[HTML]{24292F} 6}          & 0.38                             & \textbf{ 0.42  }  \\\bottomrule                
\end{tabular}
\end{table*}

\begin{table*}
\centering
\scriptsize
\caption{Success rate before and after the incorporation of prior knowledge for Ravens.}\label{tab:ravens_prior}
\begin{tabular}{lccc}\toprule
              & Ranking of  performance  by Transporter network \cite{zeng2020transporter} & CAMRL w/o  prior  success rate & CAMRL w/ prior  success rate \\\midrule
Block-insertion     & 2                                         & 94.26                          &          \textbf{94.78 }                      \\
Place-red-in-green  & 1                                         & 
93.44                          & \textbf{93.73}                           \\
Towers-of-hanoi     & 4                                         &  95.12                          & \textbf{95.34    }                       \\
Align-box-corner    & 3                                         & 
 93.43                       & \textbf{94.01    }                       \\
Stack-block-pyramid & 10                                        & \textbf{77.15 }                                  & 74.24                           \\
Palletizing-boxes   & 5                                         & 94.32                                   & \textbf{94.68 }                          \\
Assembling-kits     & 7                                         &  93.95                          & \textbf{93.99   }                        \\
Packing-boxes       & 9                                         & \textbf{79.38 }                                  & 79.06                           \\
Manipulating-rope   & 8                                         &  89.27                          & \textbf{90.24   }                        \\
Sweeping-piles      & 6                                         &  93.64                          & \textbf{93.87  }               \\\bottomrule         
\end{tabular}

\end{table*}

\begin{table*}
\centering
\scriptsize
\caption{Success rate before and after the incorporation of prior knowledge for RLBench.} \label{tab:rlbench_prior}
\begin{tabular}{lccc}
\toprule
& Ranking of  performance by \cite{liu2022auto}  & CAMRL w/o  prior success rate & CAMRL w/ prior  success rate \\
\midrule
Reach Target                & {\color[HTML]{24292F} 1}  & 99.87 & \textbf{99.93} \\
Push Button                 & 2                         & 97.24 & \textbf{98.02} \\
Pick And Lift               & {\color[HTML]{24292F} 3}  & 91.58 & \textbf{92.04} \\
Pick Up Cup                 & {\color[HTML]{24292F} 4}  & 86.48 & \textbf{86.53} \\
Put Knife on Chopping Board & {\color[HTML]{24292F} 9}  & 50.91 & \textbf{55.46 }\\
Take Money Out Safe         & {\color[HTML]{24292F} 8}  & \textbf{66.84 }& 66.79 \\
Put Money In Safe           & {\color[HTML]{24292F} 6}  & 80.38 & \textbf{82.44} \\
Pick Up Umbrella            & {\color[HTML]{24292F} 5}  & 81.49 & \textbf{82.76} \\
Stack Wine                  & {\color[HTML]{24292F} 10} & 24.59 & \textbf{26.98} \\
Slide Block To Target       & {\color[HTML]{24292F} 7}  & 79.47 & \textbf{81.46 } \\
\bottomrule
\end{tabular}

\end{table*}

\begin{figure*}[t]
    \centering
    \includegraphics[width=.8\linewidth]{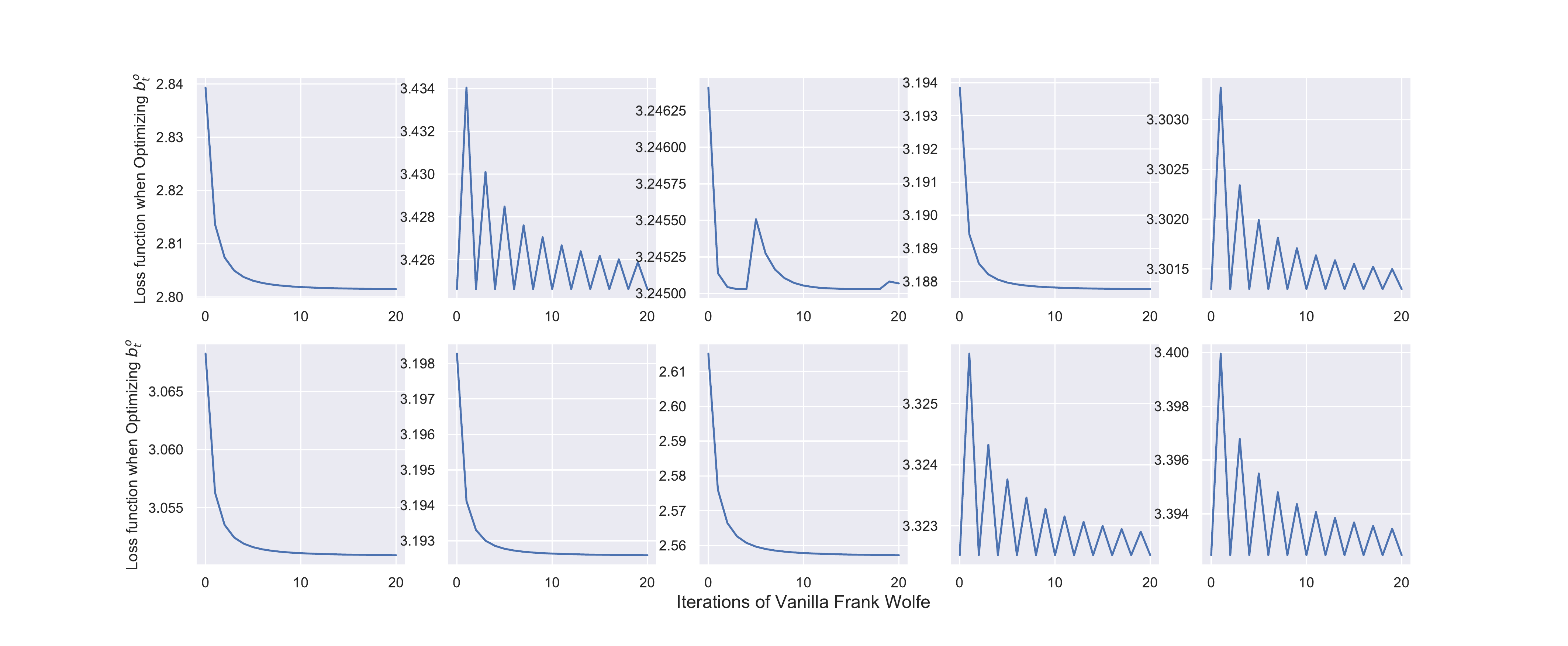}
    \caption{Convergence of the vanilla Frank-Wolfe algorithm to optimize $b_t^o$.}
    \label{fig:frank}
\end{figure*}

\subsection{Architecture}

For tasks without visual observations, we directly adopt the deep neural network as the actor network as well as the critic network. Below is the architecture of the deep neural network. 
\begin{python}
# The architecture of the actor network:

nn.Sequential(
    nn.Linear(n_o, 128),
    nn.ReLU(),
    nn.Linear(128, 64),
    nn.ReLU(),
    nn.Linear(64, 32),
    nn.ReLU(),
    nn.Linear(32, 16),
    nn.ReLU(),
    nn.Linear(16, n_a)
)

# The architecture of the critic network:

nn.Sequential(
    nn.Linear(n_o, 128),
    nn.ReLU(),
    nn.Linear(128, 64),
    nn.ReLU(),
    nn.Linear(64, 32),
    nn.ReLU(),
    nn.Linear(32, 16),
    nn.ReLU(),
    nn.Linear(16, 1)
)
\end{python}

where $n_o$ is the dimension of the observation and $n_a$ represents the number of the optional actions for tasks with discrete action space or the dimension of the action for tasks with continuous action space.

For tasks with visual observations, we perform two convolutional layers, self.conv1 and self.conv2, on the original observation, flatten the output by self.conv2, and then obtain the reshaped observation. After this, we perform the same operation as tasks
without visual observations.
\begin{python}
# in_channel, out_channel, kennel_size, stride
self.conv1 = nn.Conv2d(3, 1, (3, 4), (12,16))  
self.conv2 = nn.Conv2d(1, 1, (3, 4), (3,4))  
\end{python}

\subsection{Results visualization}

\begin{figure*}
      \centering
      \includegraphics[width=0.7\linewidth]{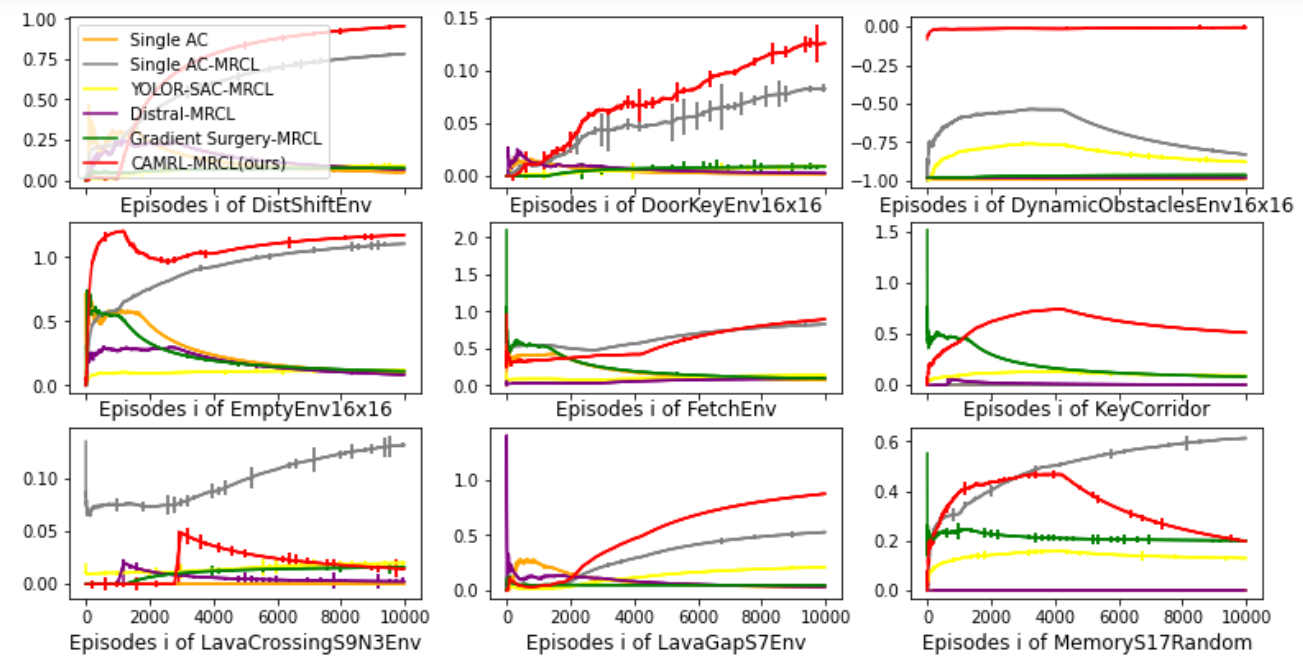} 
      \caption{Averaged reward curve of the first $10K$ episodes for Gym-minigrid tasks (each experiment repeated $5$ times).  The higher the metric, the better performance of the model.}
      \label{fig:minigrid}
      \includegraphics[width=0.7\linewidth]{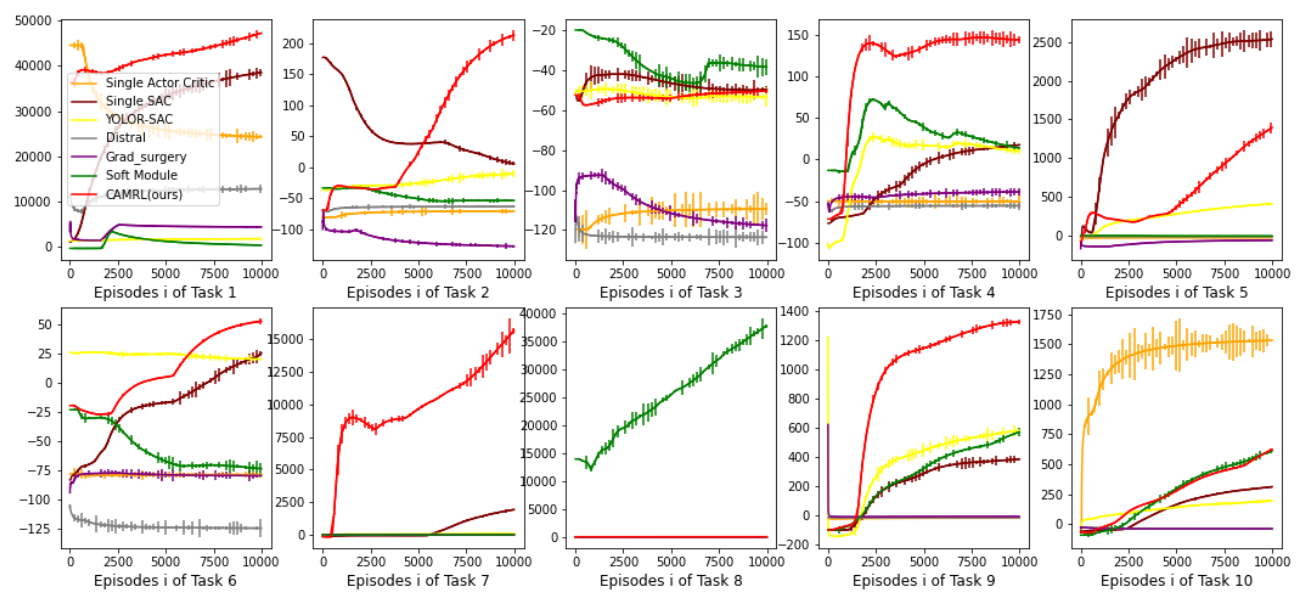}
     
      \caption{Averaged reward curve of the first $10K$ episodes for MT10 tasks (each experiment repeated $5$ times).  The higher the metric, the better performance of the model.} \label{fig:mt10}
 
     \includegraphics[width=0.3\linewidth]{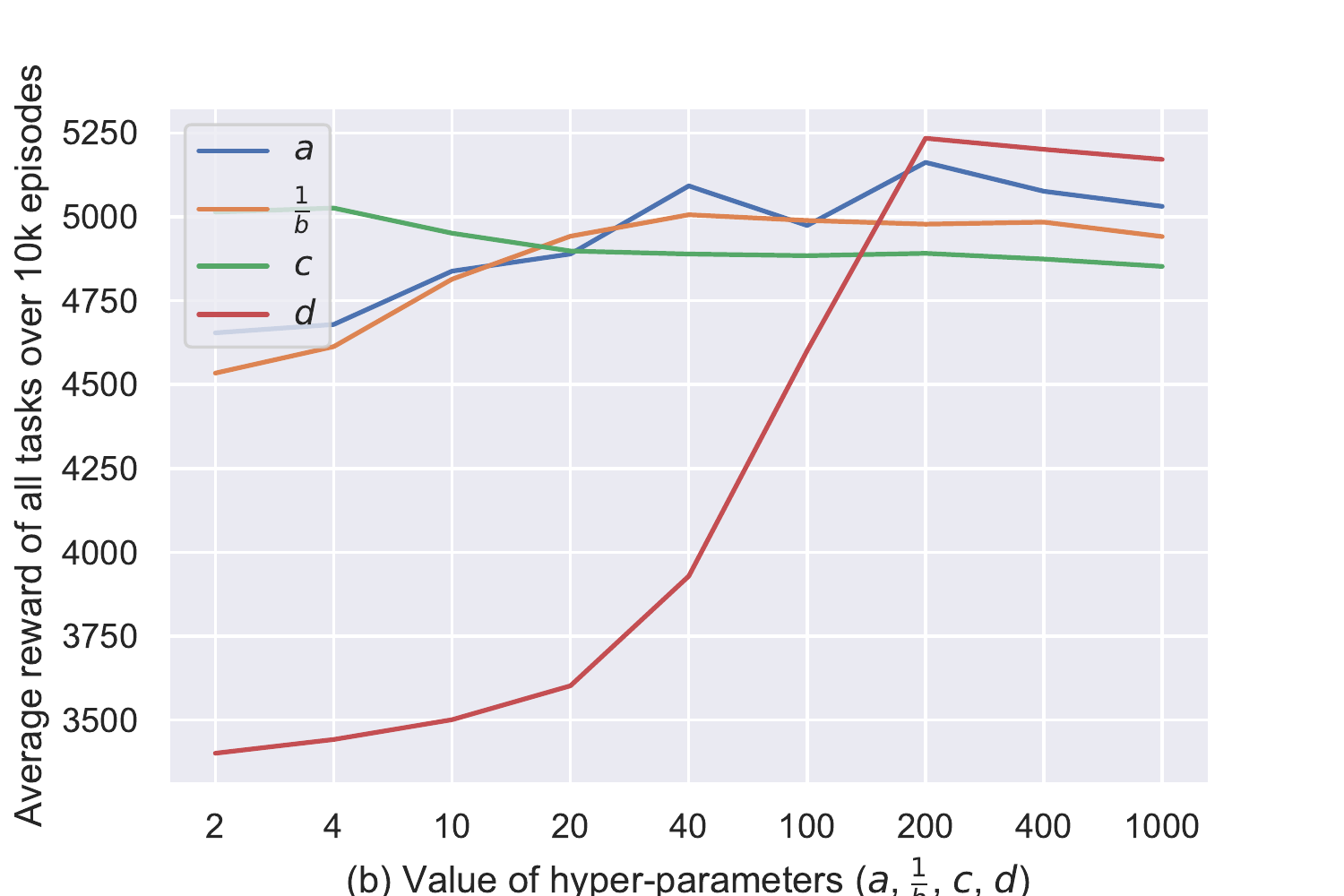}
     \caption{Hyper-parameter analysis of CAMRL.} \label{fig:para}
\end{figure*}

\begin{figure*}
      \centering
      \includegraphics[width=16cm,height=25cm]{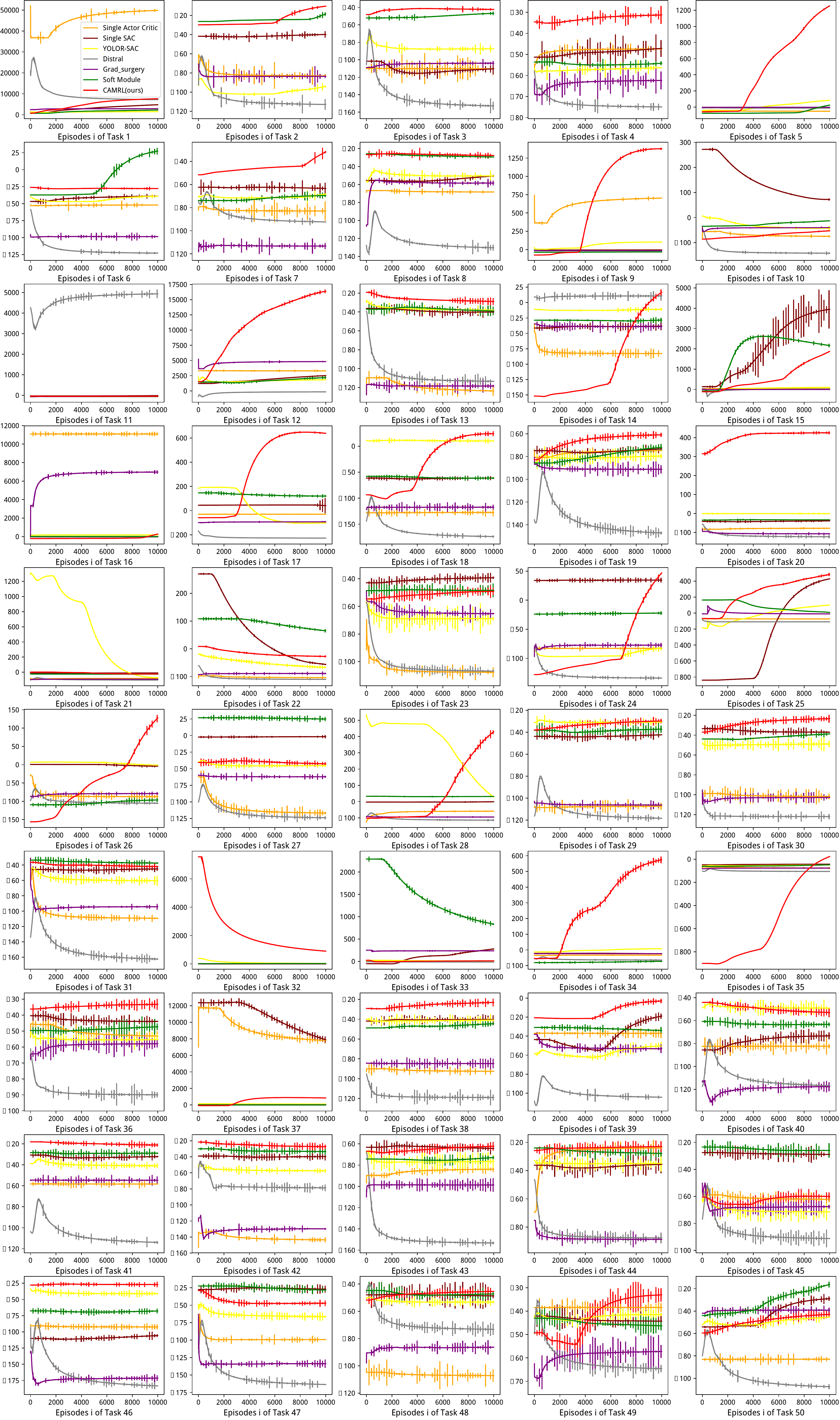}4
      \caption{Averaged reward curve of the first $10K$ episodes for MT50 tasks (each experiment repeated $5$ times).  The higher the metric, the better performance of the model.}  \label{fig:mt50}
  \end{figure*}

\begin{figure*}
      \centering
      \includegraphics[width=16cm,height=6cm]{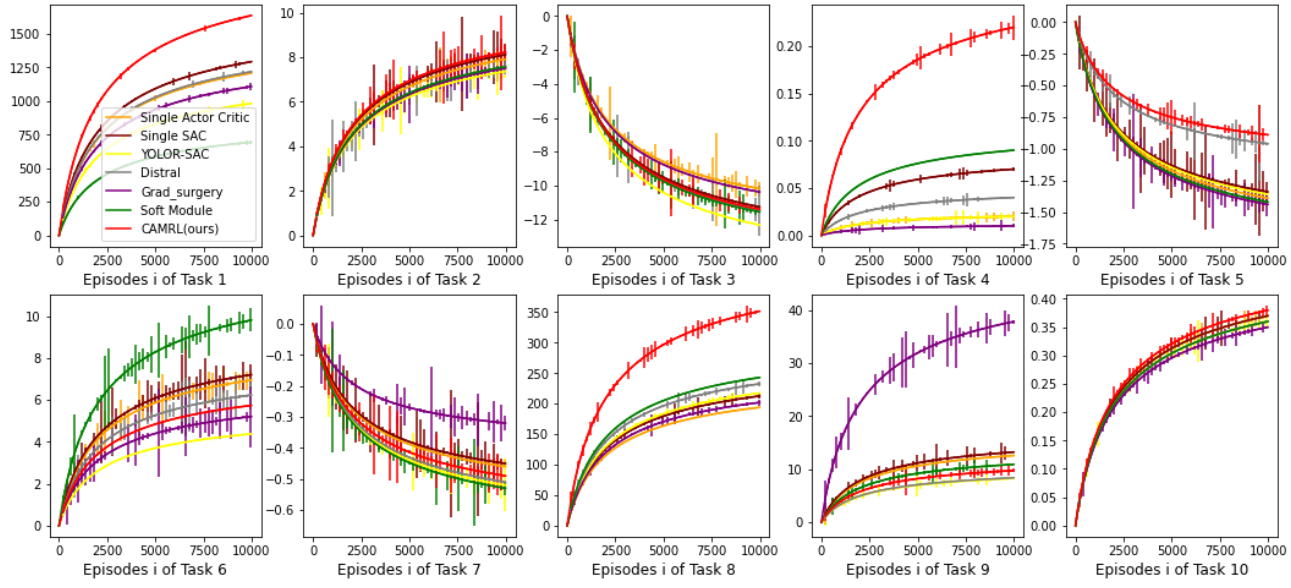}
      \caption{Averaged reward curve of the first $10K$ episodes for Atari tasks (each experiment repeated $5$ times).  The higher the metric, the better performance of the model.} \label{fig:atari}
   \vspace{1cm}\includegraphics[width=16cm,height=6cm]{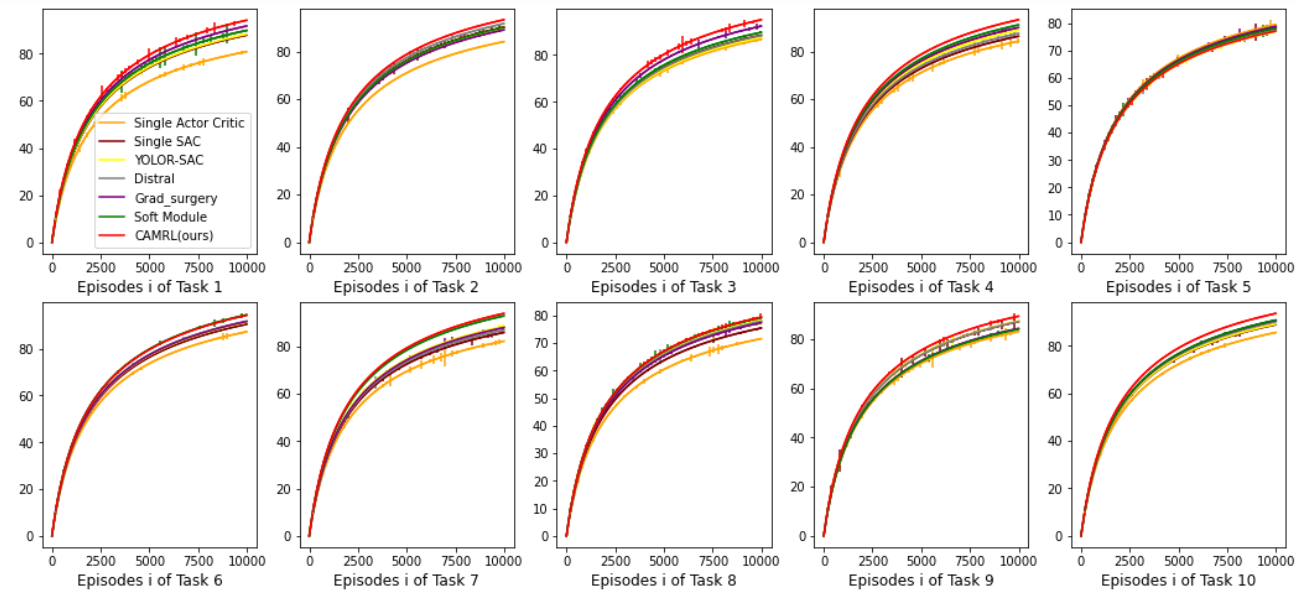}
      \caption{Averaged reward curve of the first $10K$ episodes for Ravens tasks (each experiment repeated $5$ times).  The higher the metric, the better performance of the model.} \label{fig:ravens}
  \vspace{1cm}\includegraphics[width=16cm,height=6cm]{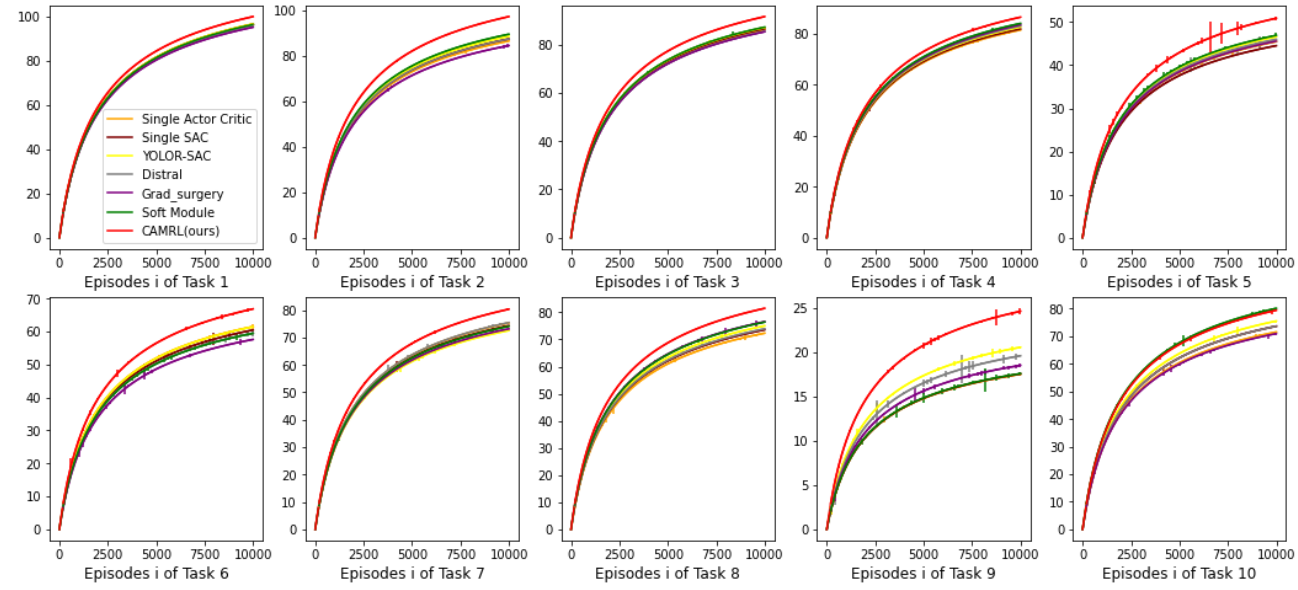}
      \caption{Averaged reward curve of the first $10K$ episodes for RLBench tasks (each experiment repeated $5$ times).  The higher the metric, the better performance of the model.}
      \label{fig:rlbench}
  \end{figure*}

We plot the performance of our selected baselines for gym-minigrid, MT10,  MT50, Atari, Ravens, and RLBench in Figures \ref{fig:minigrid}, \ref{fig:mt10},  \ref{fig:mt50}, \ref{fig:atari}, \ref{fig:ravens}, and \ref{fig:rlbench}.

\begin{table*}
\centering
\tiny 
\caption{Ablation study. Here '-' means 'without', and '+' means 'CAMRL with a specific setting'. $a$, $b$, and $c$ are the coefficients of the indicator $I_{mul}$. $\lambda_i (i\in [4])$ are coefficients of each term (except the first term) in Eq. (\ref{eq:15}).}
\label{tab:ablation study}
\begin{tabular}{lccccccccccc}
\toprule
Method \textbackslash{} Reward    & Pick and place& Pushing& Reaching& Door opening& Button press& Peg insertion side& Window opening& Window closing& Drawer opening& and Drawer closing & Avg. reward   \\
\midrule
CAMRL                                   & \textbf{45052.2} & 204.3  & -53.3  & \textbf{138.5 } & \textbf{1382.4} & \textbf{53.3} & \textbf{6764.8} & 10.3   & \textbf{1321.7} & \textbf{1512.9}  & \textbf{5638.7}         \\
AMRL                                   & 10841.9 & 10.1   & -62.2  & 30.4   & 984.2  & 2.9    & 5481.7 & \textbf{21.1}  & 104.6  & 1033.8  & 1844.9                 \\
- mode switch           & 18849.3 & 19.7   & \textbf{-50.4}  & 80.5   & 1163.8 & 30.6   & 5791.4 & 8.9    & 498.8  & 1402.1  & 2779.5                  \\
+ a=positive infinity                      & 35318.4 & 170.4  & -55.3  & 125.5  & 1303.8 & 44.1   & 6191.7 & 9.2    & 1184.2 & 1421.4  & 4571.3                  \\
+ b=positive infinity                     & 35024.8 & 149.8  & -58.9  & 104.8  & 1284.3 & 41.7   & 5938.1 & 9.1    & 872.9  & 1284.5  & 4465.1                  \\
+ c=0                         & 38529.1 & 192.6  & -53.8  & 134.9  & 1389.4 & 47.9   & 6268.8 & 10.6   & 1288.7 & 1482.9  & 4929.1                  \\
+ d=0                         & 23029.4 & 21.8   & -61.9  & 94.7   & 1228.6 & 36.1   & 5893.3 & 8.4    & 841.2  & 1465.5  & 3255.7                  \\
+ $\lambda_2$=0 & 36931.3 & 109.4  & -57.7  & 112.9  & 1255.7 & 42.6   & 4193.5 & 14.9   & 1201.8 & 1449.2  & 4525.4                  \\
+  $\lambda_3$=0 & 43093.8 & \textbf{215.4}  & -56.3  & 129.4  & 1372.5 & 47.7   & 6288.3 & 12.6   & 1279.1 & 1485.6  & 5386.8                  \\
+  $\lambda_4$=0 & 44226.7 & 202.5  & -53.8  & 130.8  & 1368.4 & 48.1   & 6631.2 & 9.3    & 1249.5 & 1474.5  &5528.7  \\     
\bottomrule
\end{tabular}
\end{table*}
\subsection{Convergence of Vanilla Frank-Wolfe Algorithm}
\subsubsection{Experimental Analysis}

We randomly select $10$ moments of performing the Vanilla Frank-Wolfe algorithm when optimizing  $b_t^o$  and 
visualize the optimization process in Figure \ref{fig:frank}. The evaluation metric is the objective function in Eq. \eqref{eq:5} with the task $t$ fixed.

\subsubsection{Theoretical Analysis}
\begin{corollary}
 Denote $x_m$ as the solution generated by vanilla Frank–Wolfe  at the $m$-th round. Then
 \begin{align}
          & \underset{1\leq j\leq m }{\min}\underset{s\geq 0,\|s\|_{1}\leq radius}{\max}\langle\nabla f(x_m),x_m-s\rangle     \label{eq:11}\\
          & \leq \frac{\max\{2h_0,4\{\lambda_0(\mu_1\!+\!\mu_2)D_1\!+\!\lambda_1 D_2^2\!+\! 2(\lambda_2\!+\!\lambda_3+\lambda_4) d T\}T \}}{\sqrt{m+1}},\nonumber
 \end{align} 
 where $h_0=f(x_0)-\underset{x\geq 0, \|x\|_{1}\leq radius}{\min} f(x)$, $D_1$ is the upper bound of $\{|\mathcal{L}(w_i)|\}_{i\in [T]}$, and $D_2$ is the $2$-norm  upper bound of the
 neural network weights. Note that, in our experiments, $D_2\leq 20$.
 
 When $f$ is  convex and differentiable, 
 \begin{align}
          & \underset{1\leq j\leq m }{\min}\underset{s\geq 0,\|s\|_{1}\leq radius}{\max}\langle\nabla f(x_m),x_m-s\rangle \nonumber\\
          &   \leq  \frac{8\{\lambda_0(\mu_1+\mu_2)D_1+\lambda_1 D_2^2+2(\lambda_2+\lambda_3+\lambda_4) d T\}T}{m+1}.
     \label{eq:12}
 \end{align} 
 
To sum up, the convergence rate of  vanilla Frank–Wolfe  on problem Eq. \eqref{eq:6} is between $\frac{1}{m}$ and $\frac{1}{\sqrt{m}}$.
 
 \end{corollary}
 
\begin{remark}
$\underset{s\geq 0,\|s\|_{1}\leq radius}{\max} \langle\nabla f(x_m),x_m-s \rangle$ is the standard evaluation metric of  vanilla Frank–Wolfe.
\end{remark}
\begin{proof}
Recall that 
  $f(b_{t}^{o})=\lambda_0[(1+\mu_1 \sum_{j\in [T]\backslash \{t\}} B_{tj})  \mathcal{L}(w_{t} ) -\mu_2 (b_t^o)^\top l_t^o ] +
\lambda_1 \sum_{s \in \mathcal{U} \backslash t}\|w_{s}-\sum_{j=1}^{i-1} B_{\pi(j) s} w_{\pi(j)}-B_{t s} w_{t}\|_{2}^{2}
+\lambda_2 \sum_{j\in [q]}(j-y{\prime}_{i_j})^2+\lambda_3 \sum_{j\in [T]}(rank^{(1)}_j-y{\prime\prime}_{j})^2 +\lambda_4 \sum_{j\in [T]}(rank^{(2)}_j-y{\prime\prime}_{j})^2$. 

Among the terms in  $f(b_{t}^{o})$,
$(1+\mu_1 \sum_{j\in [T]\backslash \{t\}} B_{tj})  \mathcal{L}(w_{t} ) -\mu_2 (b_t^o)^\top l_t^o$ is $(\mu_1+\mu_2)D_1 T$-Lipschitz  with respect to $b_t^o$ and $\lambda_1 \sum_{s \in \mathcal{U} \backslash t}\|w_{s}-\sum_{j=1}^{i-1} B_{\pi(j) s} w_{\pi(j)}-B_{t s} w_{t}\|_{2}^{2}$  is $\lambda_1 D_2^2 T$-Lipschitz  with respect to $b_t^o$, respectively.

According to $[\tanh (x)]^{\prime}=1-(\tanh (x))^2$, it can be easily deduced that $\lambda_2 \sum_{j\in [q]}(j-y{\prime}_{i_j})^2+\lambda_3 \sum_{j\in [T]}(rank_j-y{\prime\prime}_{j})^2+\lambda_4 \sum_{j\in [T]}(rank^{(2)}_j-y{\prime\prime}_{j})^2$ is $2(\lambda_2+\lambda_3+\lambda_4) dT^2$-Lipschitz  with respect to $b_t^o$.

By integrating the above terms, we obtain  the $\{\lambda_0 (\mu_1+\mu_2)D_1+\lambda_1 D_2^2+2(\lambda_2+\lambda_3+\lambda_4) dT\}T$-Lipschitz property of $f(b_t^o)$. Also note that the diameter of $b_t^o$'s feasible solution space is no greater than $1$, so by  Equation 3 in  \cite{Lacoste}  and Equation 3.11 in \cite{Nesterov18}  we can easily deduce 
 the  two inequalities in Eq. \eqref{eq:11} and Eq. \eqref{eq:12}.
\end{proof}

\end{document}